\documentclass[letterpaper]{article}

\PassOptionsToPackage{numbers, square}{natbib}

  \usepackage[final]{neurips_2020}

\usepackage[utf8]{inputenc} 
\usepackage[T1]{fontenc}    
\usepackage{hyperref}       
\usepackage{url}            
\usepackage{booktabs}       
\usepackage{amsfonts}       
\usepackage{nicefrac}       
\usepackage{microtype}      
\usepackage{graphicx,url}
\usepackage{subcaption}
\usepackage{multicol}
\usepackage{amsmath}
\usepackage{amssymb}
\usepackage{hyperref}
\usepackage{algorithmic}
\usepackage{algorithm}
\hypersetup{colorlinks,linkcolor={blue},citecolor={blue},urlcolor={blue}}
\renewcommand{\algorithmiccomment}[1]{\bgroup\hfill$\triangleright$~#1\egroup}
\newcommand{\MiniFigureWidth}{0.19}
\newcommand{\MnistFigureWidth}{0.16}
\DeclareMathOperator*{\argmax}{argmax} 
\DeclareMathOperator*{\argmin}{argmin} 

\title{Feature Importance Ranking for Deep Learning}
\author{%
 \hspace*{-6mm} Maksymilian A.~Wojtas ~~~~~~~~ Ke Chen\\
  Department of Computer Science,
  The University of Manchester, Manchester M13 9PL, U.K. \\
  \texttt{\{maksymilian.wojtas,ke.chen\}@manchester.ac.uk} \\
}

\begin{document}

\maketitle

\begin{abstract}
Feature importance ranking has become a powerful tool for explainable AI. However, its nature of combinatorial optimization poses a great challenge for deep learning. In this paper, we propose a novel dual-net architecture consisting of operator and selector for discovery of an optimal feature subset of a fixed size and ranking the importance of those features in the optimal subset simultaneously. During learning, the operator is trained for a supervised learning task via optimal feature subset candidates generated by the selector that learns predicting the learning performance of the operator working on different optimal subset candidates. We develop an alternate learning algorithm that trains two nets jointly and incorporates a stochastic local search procedure into learning to address the combinatorial optimization challenge. In deployment, the selector generates an optimal feature subset and ranks feature importance, while the operator makes predictions based on the optimal subset for test data.
A thorough evaluation on synthetic, benchmark and real data sets suggests that our approach outperforms several state-of-the-art feature importance ranking and supervised feature selection methods.
(Our source code is available:
\url{https://github.com/maksym33/FeatureImportanceDL})
\end{abstract}

\section{Introduction}
\label{sect:intro}

In machine learning, feature importance ranking (FIR) refers to a task that measures contributions of individual input features (variables) to the performance of a supervised learning model. FIR has become one of powerful tools in explainable/interpretable AI \cite{samek2017explainable} to facilitate understanding of decision-making by a learning system and discovery of key factors in a specific domain, e.g., in medicine, what genes are likely main causes of a cancer \cite{guyon2002RFE}.

Due to the existence of correlated/dependent and irrelevant features to targets in high-dimensional real data, feature selection \cite{guyon2003SFintro} is often employed to address the well-known curse of dimensionality challenge and to improve the generalization of a learning system, where a subset of optimal features is selected in terms of the pre-defined criteria to maximize the performance of a learning system. Feature selection may be conducted at either population or instance level; the populationwise methods would find out an optimal feature subset collectively for all the instances in a population, while the instancewise ones tend to uncover a subset of salient features specific to a single instance. In practice, FIR is always closely associated with feature selection by ranking the importance of those features in an optimal subset and can also be used as a proxy for feature selection, e.g., \cite{guyon2002RFE,song2007BAHSIC1,song2012BAHSIC2}.

Deep learning has turned out to be extremely powerful in intelligent system development but its purported ``black box'' nature makes it extremely difficult to be applied to tasks demanding explainability/interpretability. Recently, FIR for deep learning has become an active research area where most works focus on instancewise FIR \cite{hooker2019benchmark} and only few works exist for populationwise FIR/feature selection, e.g., \cite{Li2016DFS}. In a populationwise scenario, feature selection needs to find an optimum in detecting any functional dependence between input data and targets, which is NP-hard in general  \cite{weston2003NP-hard-FS}. High degree of nonlinearity in deep learning execrates this combinatorial optimization problem.

In this paper, we address a populationwise FIR issue in deep learning: for a feature set, finding an optimal feature subset of a fixed size that maximizes the performance of a deep neural network and ranking the importance of all the features in this optimal subset simultaneously. To tackle this problem, we propose a novel dual-net neural architecture, where an operator net works for a supervised learning task via optimal subset candidates provided by a selector net that learns finding the optimal feature subset and ranking feature importance via the learning performance feedback of the operator. Two nets are jointly trained in an alternate manner. After learning, the selector net is used to find an optimal feature subset and rank feature importance, while the operator net makes predictions based on the optimal feature subset for test data.
A thorough evaluation on synthetic, benchmark and real datasets via a comparative study manifests that our approach leveraged by deep learning outperforms several state-of-the-art FIR and supervised feature selection methods.

\section{Related Work}
\label{sect:related}

In the context of deep learning, there exist three methods for FIR; i.e., \emph{regularization}, \emph{greedy search} and \emph{averaged input gradient}.
The deep feature selection (DFS) \cite{Li2016DFS} was proposed for FIR with the same idea behind the regularized linear models \cite{tibshirani1996LASSO,zou2005Elasticnet}.
The DFS suffers from several issues, e.g. a high computational burden in finding an optimal regularization hyper-parameters and vanishing gradient. Moreover, the weight-shrinkage idea \cite{tibshirani1996LASSO,zou2005Elasticnet} may not always work for complex dependence between input features and targets since the use of shrunk weights as feature importance is theoretically justifiable to linear models only. It seems straightforward to apply a greedy search method, e.g., forward subset selection (FS) \cite{friedman2001els}, to deep learning for FIR. Obviously, this method inevitably incurs extremely high computational cost and may end up with only a sub-optimal result. Finally, some instancewise FIR methods have been converted into populationwise ones, e.g., the averaged input gradient (AvGrad) \cite{hechtlinger2016Inputgradient} that uses the average of all the saliency maps extracted from individual instances for FIR and global aggregation \cite{ribeiro2016should,linden2019explanationbb,skrlj2020fiesan} that uses different aggregation mechanisms to achieve the populationwise feature importance ranking. As local explanations are specific at the instance level and often inconsistent with global explanations at the population level, the simple accumulation of instancewise FIR results may not work on populationwise FIR. In contrast, our method would overcome all the limitations stated above.

In machine learning,  regularized linear models, e.g., LASSO \cite{tibshirani1996LASSO}, and random forest (RF) \cite{breiman2001RF} are two off-the-shelf FIR methods. Other strong FIR methods include the SVM-based RFE \cite{guyon2002RFE} and the dependence-maximization based BAHSIC \cite{song2007BAHSIC1,song2012BAHSIC2}. In general, such methods may have the limited learning capacity for complex tasks in comparison to deep learning, and may not always work
for complex dependence between input features and targets. On the other hand, according to the definition in \cite{song2007BAHSIC1,chen2017CCM}, our FIR problem formulation can be treated a sub-problem of supervised feature selection when the size of an optimal feature subset is pre-specified. To this end, our method is closely related to several strong feature selection methods with the same setting, including those working on mutual information criteria, e.g., mRMR \cite{peng2005mRMR} and the kernel-based CCM \cite{chen2017CCM} although such methods do not consider FIR. Leveraged with deep learning, our approach is more effective than those aforementioned FIR and supervised feature selection methods, as manifested in our experiments.

\section{Method}
\label{sect:method}

\subsection{Problem Formulation}
\label{subsect:problem}

Suppose
${\cal D} = \{ {\cal X}, {\cal Y} \}$
is a dataset used for supervised learning. In this data set,  $(\pmb{x}, \pmb{y})$  is a training example, where $\pmb{x} \in {\cal X}$ is a vector of $d$ features and $\pmb{y} \in {\cal Y}$ is its corresponding target. Let ${\pmb m} \in {\cal M}$ denote a $d$-dimensional binary mask vector of $0/1$ elements, where $||{\pmb m}||_0 = s$, $s<d$ and $|{\cal M}|= \binom{d}{s}$. Thus, we can use such a mask vector to indicate a feature subset: $\{{\pmb x} \otimes {\pmb m} \}_{{\pmb x} \in {\cal X}}$, where $\otimes$ denotes Hadamard product that yield a subset of $s$ features for any instance ${\pmb x} \in {\cal X}$.
Assume that ${\cal Q}({\pmb x}, {\pmb m})$ quantifies the instance-level performance of a learning system trained on ${\cal D}$ via a feature subset, $\{{\pmb x} \otimes {\pmb m} \}_{{\pmb x} \in {\cal X}}$, the \emph{feature importance ranking} (FIR) can then be formulated as follows:
\begin{equation}
\big ( {\pmb m}^*,{\rm Score}({\pmb m}^*) \big ) = \argmax_{{\pmb m} \in {\cal M}}
\sum_{ {\pmb x} \in {\cal X}} {\cal Q}({\pmb x}, {\pmb m}),
\label{eq:fir-def}
\end{equation}
where ${\pmb m}^*$ is the indicator of an optimal feature subset discovered by an FIR algorithm and ${\rm Score}({\pmb m}^*)$ quantifies the importance of all the selected features in this optimal subset.

\begin{figure*}[t!]
    \centering
    \begin{subfigure}{0.42\textwidth}
    \hspace*{-3mm}
        \centering
        \fbox{\includegraphics[width=\textwidth]{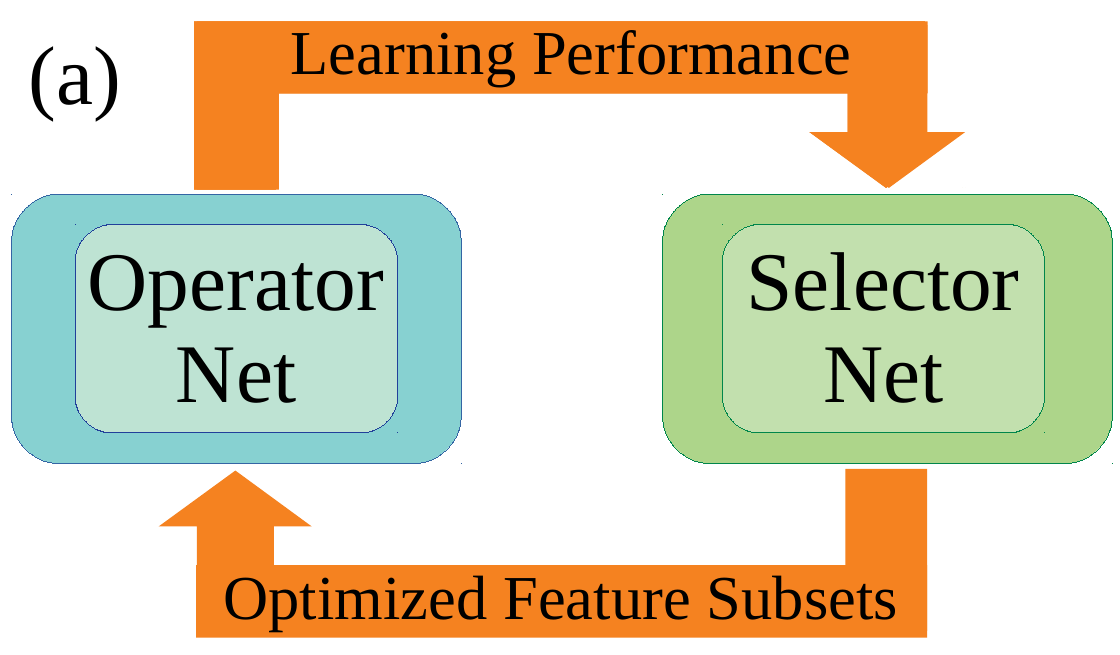}}
    \end{subfigure}
    \begin{subfigure}{0.549\textwidth}
        \centering
        \fbox{\includegraphics[width=\textwidth]{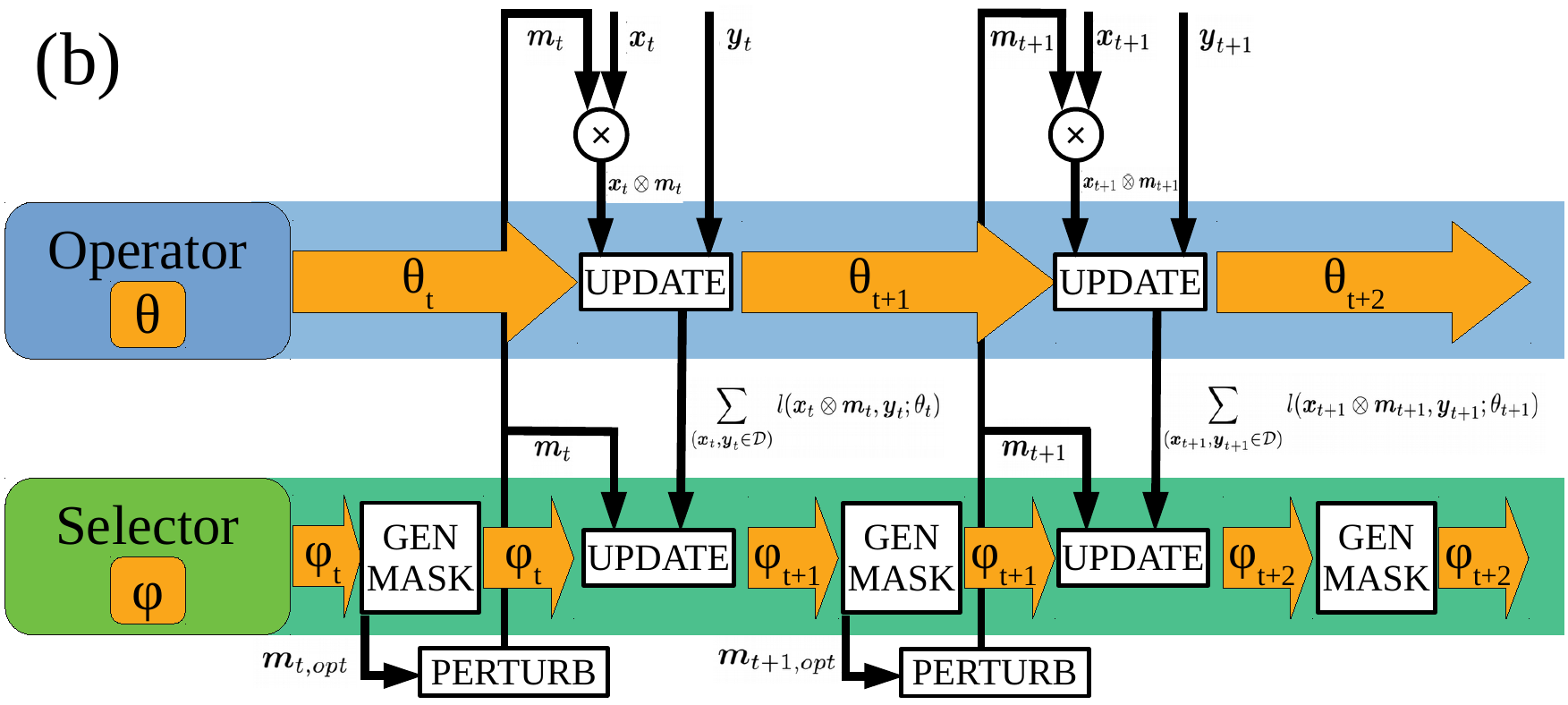}}
    \end{subfigure}
    \caption{Our feature importance ranking model. (a) Dual-net architecture. (b) Parameter update.}
    \label{fig:model}
\end{figure*}

Ideally, an FIR approach should be able to: 1) detect any functional dependence between input features and targets;
2) rank the importance of all the selected features to reflect their contributions to the learning performance; and 3) preserve the detected functional dependence and the feature importance ranking in test data.

\subsection{Model Description}
\label{subsect:model}

To tackle the FIR problem stated in Eq.(\ref{eq:fir-def}) effectively with three criteria described in Sect. \ref{subsect:problem}, we propose a deep learning model of dual nets, \emph{operator} and \emph{selector}, as shown in Fig. \ref{fig:model}(a). The operator net is employed to accomplish a supervised learning task, e.g., classification or regression, on a given feature subset provided by the selector net, while the selector net is designated to learn finding out an optimal feature subset based on the performance feedback of the operator net working on optimal feature subset candidates during learning. Both the operator and the selector nets are trained jointly in an alternate manner (c.f. Sect. \ref{subsect:learn}) to reach a synergy for the FIR.

Technically, the operator is carried out with a deep neural network parameterized with $\theta$, $f_O(\theta; \pmb x, \pmb m)$, for a given task, e.g., multi-layer perceptron (MLP) or convolution neural network (CNN). This net is trained on ${\cal D}$ based on different feature subsets to learn
$f_O\!: {\cal X} \times {\cal M} \rightarrow {\cal Y}$.
After learning (c.f. Sect. \ref{subsect:learn}), the trained operator net,
$f_O(\theta^*; \pmb x, {\pmb m}^*)$, is applied to the test data for prediction, where $\theta^*$ is the optimal parameters of the operator net and $\pmb m^*$ is generated by the trained selector net (c.f. Sect. \ref{subsect:deployment}).

In our method, the selector is implemented with an MLP parameterized with $\varphi$, $f_S(\varphi; \pmb m)$. As defined in Eq.(\ref{eq:fir-def}), a selected optimal feature subset should maximize the averaging  performance of the operator quantified by ${\cal Q}({\pmb x}, {\pmb m})$ for all $\pmb x \in \cal X$. Thus, we want the selector net to learn predicting the averaging performance of the operator net on different feature subsets; i.e., $f_S\!\!: {\cal M} \rightarrow \mathbb{R}$. After being trained properly (c.f. Sect. \ref{subsect:learn}), we can use an algorithm working on the trained selector net of the optimal parameters $\varphi^*$, $f_S(\varphi^*; \pmb m)$, to generate an optimal feature subset indicated by ${\pmb m}^*$ and rank feature importance to achieve ${\rm Score}({\pmb m}^*)$ (c.f. Sect. \ref{subsect:deployment}).
%

\subsection{Learning Algorithm}
\label{subsect:learn}

In essence, the FIR defined in Eq.(\ref{eq:fir-def}) is a combinatorial optimization problem. According to the no free lunch theory for optimization \cite{wolpert1997NFL},
no algorithm can perform better than a random strategy in expectation in the setting of combinatorial optimization. Therefore, our learning algorithm is developed by leveraging learning with a stochastic local search procedure enhanced by injecting noise \cite{mengshoel2008noise} on a small number of candidate feature subsets, ${\cal M}' \subset {\cal M}$, to avoid the exhaustive search.

For a training data set, ${\cal D} =  \{\cal X, \cal Y  \} = \big \{ (\pmb x, \pmb y) \big \}_{\pmb x \in {\cal X}, \pmb y \in {\cal Y}}$, a mask subset, ${\cal M}'$, converts each training example $(\pmb x, \pmb y) \in {\cal D}$ into $|{\cal M}'|$ examples:
$\big \{(\pmb x \otimes \pmb m, \pmb y) \big \}_{\pmb m \in {\cal M}'}$. Thus, the loss functions on ${\cal M}'$ (changing during learning) for the operator and the selector nets are defined respectively as follows:

\begin{subequations}
\begin{equation}
{\cal L}_O ({\cal D}, {\cal M}'; \theta) = \frac 1 {|{\cal M}'||{\cal D}|}
\sum_{\pmb m \in {\cal M}'} \sum_{(\pmb x, \pmb y) \in {\cal D}}  l(\pmb x \otimes \pmb m, \pmb y; \theta),
\label{eq:oloss}
\end{equation}
\begin{equation}
{\cal L}_S ({\cal M}'; \varphi ) =  \frac 1 {2|{\cal M}'|} \sum_{\pmb m \in {\cal M}'}
\Big ( f_S(\varphi; \pmb m) - \frac 1 {|{\cal D}|} \sum_{({\pmb x}, {\pmb y}) \in {\cal D}}
l(\pmb x \otimes \pmb m, \pmb y; \theta) \Big )^2.
\label{eq:sloss}
\end{equation}
\end{subequations}
Here, $l(\pmb x \otimes \pmb m, \pmb y; \theta)$ is an instance-level cross-entropy/categorical cross-entropy loss for binary/mutli-class classification or the mean square error (MSE) loss for regression. In Eq.(\ref{eq:sloss}), we utilize the loss of the operator net, $l(\pmb x \otimes \pmb m, \pmb y; \theta)$, to characterize its learning performance, ${\cal Q}({\pmb x}, {\pmb m})$, since maximizing ${\cal Q}({\pmb x}, {\pmb m})$ is equivalent to  minimizing $l(\pmb x \otimes \pmb m, \pmb y; \theta)$.
As described in Sect. \ref{subsect:model}, during learning, the operator net relies on the selector net to provide an optimal subset of marks, ${\cal M}'$, indicating different optimal feature subset candidates, while the selector net requires the performance feedback from the operator net, $l(\pmb x \otimes \pmb m, \pmb y; \theta)$ for all
$\pmb m \in {\cal M}'$. Two nets in our learning model hence have to be trained alternately. Below, we present the main learning steps in our learning algorithm of two phases, while the pseudo code can be found from Sect. D in supplementary materials.

\paragraph{Phase I: Initial Operator Learning via Exploration.}
From the scratch, we start training the operator net by using a small number of random feature subsets for several epochs until it can yield the different performance on different feature subsets stably. Technically, in each epoch, we randomly draw a subset of different masks, ${\cal M}_1'$, from $\cal M$; i.e., ${\cal M}_1'= \big \{ {\pmb m}_i| {\pmb m}_i = {\rm Random}({\cal M}, s) \big \}_{i=1}^{|{\cal M}'|}$, where ${\rm Random}({\cal M}, s)$ is a function that randomly draws a $d$-dimensional mask of $s$ one-elements and $d\!-\!s$ zero-elements from $\cal M$. If $\theta$ is trained by stochastic gradient decent (SGD), then it is updated by
$\theta'' \triangleq \theta' - \eta \nabla_{\theta}
{\cal L}_O ({\cal D}, {\cal M}_1'; \theta)|_{\theta=\theta'}$\footnote{Parameters are actually updated on a batch $\cal B$ randomly drawn from $\cal D$, hence $\frac {|\cal D|} {|\cal B|}$ times in an epoch.} where $\eta$ is a learning rate. After $E_1$ epochs, we set $\theta_1 = \theta''(E_1)$  and ${\pmb m}_{1,opt}' = \argmin_{{\pmb m} \in {\cal M}_1'} \sum_{(\pmb x, \pmb y) \in {\cal D}}  l(\pmb x \otimes \pmb m, \pmb y; \theta_1)$ to be used at the beginning of Phase II-A;
i.e., $t=1$ as shown in Fig. \ref{fig:model}(b).

\paragraph{Phase II-A: Selector Learning via Operator's Feedback.} As illustrated in Fig. \ref{fig:model}(b), the operator provides training examples for the selector at step $t$: $\big \{ \big ({\pmb m}, \frac 1 {|{\cal D}|} \sum_{({\pmb x}, {\pmb y}) \in {\cal D}} l(\pmb x \otimes \pmb m, \pmb y; \theta_t) \big ) \big \}_{{\pmb m} \in {\cal M}_t'}$. By using the SGD with initializing $\varphi_1$ randomly, the parameters in the selector net, $\varphi$, are updated by
$\varphi_{t+1} \triangleq \varphi_t - \eta \nabla_{\varphi}
{\cal L}_S ({\cal M}_t'; \varphi)|_{\varphi=\varphi_t}$. Then, we adopt an \emph{exploration-exploitation} strategy  to generate a new mask subset, ${\cal M}_{t+1}'$, for the operator learning at step $t\!+\!1$. Thus, ${\cal M}_{t+1}'$ is divided into two mutually exclusive subsets: ${\cal M}_{t+1}'={\cal M}_{t+1,1}' \cup {\cal M}_{t+1,2}'$. Motivated by the role of noise in stochastic local search \cite{mengshoel2008noise}, ${\cal M}_{t+1,1}'$ is generated via exploration to avoid overfitting:
${\cal M}_{t+1,1}' = \big \{ {\pmb m}_i| {\pmb m}_i = {\rm Random}({\cal M}, s) \big \}_{i=1}^{|{\cal M}_{t+1,1}'|}$. Motivated by the input gradient idea
 \cite{hechtlinger2016Inputgradient}, ${\cal M}_{t+1,2}'$ is generated by exploitation of the selector net, $f_S(\varphi_{t+1}; \pmb m)$, as follows:
 \textbf{a) Generation of an optimal subset}.
 Starting with  $d$-dimensional ${\pmb m}_0= \big (\frac 1 2, \cdots, \frac 1 2 \big )$, meaning that every feature has the equal chance to be selected, we have ${\pmb \delta}_{{\pmb m}_0} =
 {\frac {\partial f_S(\varphi_{t+1}; \pmb m)} {\partial {\pmb m}}}  |_{{\pmb m}={\pmb m}_0}$. As input features of the larger gradients contribute more to the learning performance of the operator, we can find top $s$ features based on their gradients by
 $({\pmb m}_{opt}, \bar{\pmb m}_{opt}) = {\rm argsort}({\pmb \delta}_{{\pmb m}_0}, s)$ where ${\pmb m}_{opt}$ is the mask  to indicate top $s$ features and $\bar{\pmb m}_{opt}$ is the mask for the remaining $d\!-\!s$ features. To ensure the optimality of ${\pmb m}_{opt}$, we come up with a three-step \emph{validation} procedure: \textbf{i}) Re-evaluate the contributions of top $s$ features by $({\pmb m}_{opt}, \bar{\pmb m}_{opt}) = {\rm argsort}({\pmb \delta}_{{\pmb m}_{opt}}, s)$ where
 ${\pmb \delta}_{{\pmb m}_{opt}} =
 {\frac {\partial f_S(\varphi_{t+1}; \pmb m)} {\partial {\pmb m}}} |_{{\pmb m}={\pmb m}_{opt}}$; \textbf{ii}) Replace a feature of negative gradient in ${\pmb m}_{opt}$ with the one of the largest gradient in $\bar{\pmb m}_{opt}$ if there exists;
 \textbf{iii}) Further check the optimality via a function
 $({\pmb m}_{opt}', \bar{\pmb m}_{opt}')={\rm swap}({\pmb m}_{opt}, \bar{\pmb m}_{opt})$ that yields ${\pmb m}_{opt}'$ by swapping between the feature of least gradient in ${\pmb m}_{opt}$ and the one of the largest gradient in $\bar{\pmb m}_{opt}$. Repeat (i)-(iii) until  $f_S(\varphi_{t+1}; {\pmb m}_{opt}) \leq f_S(\varphi_{t+1}; {\pmb m}_{opt}')$. After going through the validation procedure,  ${\pmb m}_{t+1,opt}$ is obtained for step $t\!+\!1$.
 \textbf{b) Generation of optimal subset candidates via perturbation}. As the optimal subset ${\pmb m}_{t+1,opt}$ might be a local optimum, we would further inject noise to generate more optimal subset candidates by a perturbation function ${\rm Perturb}({\pmb m}_{opt},s_p)$. For $s_p < s$, ${\rm Perturb}({\pmb m}_{opt},s_p)$ randomly flips $s_p$ different elements in
 ${\pmb m}_{opt}$/$\bar{\pmb m}_{opt}$ from 1/0 to 0/1 and swaps between changed elements in ${\pmb m}_{opt}$ and $\bar{\pmb m}_{opt}$.
 Applying ${\rm Perturb}({\pmb m}_{opt},s_p)$ repeatedly leads to multiple optimal subset candidates;
 \textbf{c) Formation of optimal subset candidates}. Assembling \textbf{a)} and \textbf{b)} leads to ${\cal M}_{t+1,2}' = \{{\pmb m}_{t,best}\} \cup
 \{{\pmb m}_{t+1,opt}\} \cup \big \{{\pmb m}_i|{\pmb m}_i={\rm Perturb}({\pmb m}_{t+1,opt},s_p) \big \}_{i=1}^{|{\cal M}_{t+1,2}'|-2}$. Here, we always include ${\pmb m}_{t,best}$, the subset that leads to the best learning performance of the operator net in the last step (step $t$), as the most important subset candidate in the current step (step $t\!+\!1$) in order to make the operator learning progress steadily.
 Note that ${\pmb m}_{t,best}$ may not be ${\pmb m}_{t,opt}$.


\paragraph{Phase II-B: Operator Learning via Optimal Subset Candidates from Selector.} After completing the training of Phase II-A at step $t$ , the selector net provides the optimal subset candidates, ${\cal M}_{t+1}'={\cal M}_{t+1,1}' \cup {\cal M}_{t+1,2}'$, for the operator net,
as illustrated in Fig. \ref{fig:model}(b). At step $t\!+\!1$, the operator net is thus trained based on ${\cal M}_{t+1}'$ with SGD:
$\theta_{t+1} \triangleq \theta_t - \eta \nabla_{\theta}
{\cal L}_O ({\cal D}, {\cal M}_{t+1}'; \theta)|_{\theta=\theta_t}$.

As shown in Fig. \ref{fig:model}, our alternate algorithm enables the operator and the selector nets to be trained jointly in Phase II until a pre-specified stopping condition is satisfied.

\subsection{Deployment}
\label{subsect:deployment}

After the learning described in Sect. \ref{subsect:learn} is accomplished, we obtain the optimal parameters of the operator and the selector nets, $\theta^*$ and $\varphi^*$.

By using the trained selector net, $f_S(\varphi^*; \pmb m)$, we find out an optimal feature subset with the same procedure used in Phase II-A as follows: 1) starting
with ${\pmb m}_0= \big (\frac 1 2, \cdots, \frac 1 2 \big )$, calculate the gradient ${\pmb \delta}_{{\pmb m}_0} =
 {\frac {\partial f_S(\varphi^*; \pmb m)} {\partial {\pmb m}}}  |_{{\pmb m}={\pmb m}_0}$;
2) finding top $s$ features by
 $({\pmb m}^*, \bar{\pmb m}^*) = {\rm argsort}({\pmb \delta}_{{\pmb m}_0}, s)$, where ${\pmb m}^*$ indicates the optimal subset of top $s$ features;
 and 3) going through the validation procedure described in Phase II.A to ensure the optimality of ${\pmb m}^*$. Thus, feature importance ranking on the final ${\pmb m}^*$ is done by setting ${\rm Score}({\pmb m}^*)= {\frac {\partial f_S(\varphi^*;  \pmb m)} {\partial {\pmb m}}} |_{{\pmb m}={\pmb m}^*}$ and sorting the input gradients of selected features.

During test, for a test instance, $\hat{\pmb x}$, the trained operator net, $f_O(\theta^*; \pmb x, {\pmb m})$, can be used to make a prediction, $f_O(\theta^*; \hat{\pmb x}, {\pmb m}^*)$, via $\hat{\pmb x} \otimes {\pmb m}^*$, which allows a supervised learning task to be done based on only the optimal feature subset, ${\pmb m}^*$, found out with our proposed approach.


\section{Experiments}
\label{sect:experiment}

In this section, we evaluate our approach on synthetic, benchmark and real-world datasets where we always use 5-fold cross-validation for evaluation and report the performance statistics, i.e., \emph{mean} and \emph{standard deviation} estimated on 5 folds. We describe our main settings briefly in the main text, and the details of all the experimental settings can be found from Sect. A in Supplementary Materials.

\subsection{Synthetic Data}
\label{sect:synthetic}

Our first evaluation employs 3 synthetic datasets in literature \cite{chen2017CCM,friedman2001els} for feature selection regarding regression and binary/multiclass classification as follows:

\noindent
\textbf{XOR as 4-way classification} \cite{chen2017CCM}.
Group 8 corners of the cube, $(v_0,v_1,v_2) \in \{-1,+1\}^3$, by the tuples $(v_0v_2, v_1v_2)$, leading to 4 sets of vectors paired with their negations $\{v^{(c)}, -v^{(c)}\}$. For a class $c$, a point is generated from the mixture distribution: $\frac 1 2 [N(v^{(c)}, 0.5I_3) + N(-v^{(c)}, 0.5I_3)]$. Then, form a 10-D feature vector for each example by adding 7 standard noise features, $(X_3,\cdots,X_9) \sim N(0,I_7)$.

\noindent
\textbf{Nonlinear regression} \cite{chen2017CCM}.
$Y= -2\sin(2X_0)+\max(X_1,0) + X_2 + \exp(-X_3)+ \epsilon$, where
$(X_0,\cdots,X_9) \sim N(0, I_{10})$ and $\epsilon \sim N(0,1)$, leading to a 10-D feature vector for each example.

\noindent
\textbf{Binary classification} \cite{friedman2001els}. To generate examples, set $Y=-1$ when $(X_0,\cdots,X_9) \sim N(0,I_{10})$ and $Y=+1$ when $X_0$ through $X_3$ are standard normal conditioned on $9 \leq \sum_{i=0}^3 X_i^2 \leq 16$ and $(X_4,\cdots,X_9) \sim N(0,I_6)$, resulting in a 10-D feature vector for each example.

For each dataset, we randomly generate 512 and 1024 examples, respectively,  for training and test.
With our problem formulation described in Sect. \ref{subsect:problem}, our experiment on synthetic data simulates an application scenario that selects $s$ out of $d$ features where $s$ is larger than the number of features relevant to the target in a dataset. As there are up to 4 relevant features in the above 3 datasets, we choose $s=5$ in our experiment and compare with all the methods reviewed in Sect. \ref{sect:related}, including DFS \cite{Li2016DFS}, AvGrad \cite{hechtlinger2016Inputgradient}, FS \cite{friedman2001els} based on MLP, LASSO \cite{tibshirani1996LASSO}, RF \cite{breiman2001RF},  RFE \cite{guyon2002RFE}, BAHSIC \cite{song2007BAHSIC1,song2012BAHSIC2}, mRMR \cite{peng2005mRMR} and CCM \cite{chen2017CCM}. According to a taxonomy \cite{guyon2003SFintro}, DFS, AvGrad, RF and ours are embedding methods, FS is a wrapper method and all the others are filtering methods. For those filtering methods, we use the exactly same kernel SVM/SVR described in those papers  \cite{guyon2002RFE,song2007BAHSIC1,peng2005mRMR,song2012BAHSIC2,chen2017CCM} and an MLP on LASSO for classification/regression.
While DFS, AvGrad, LASSO and RF work on FIR for all 10 features, all other methods work with the same setting as ours by finding out top 5 features and FIR.

\begin{figure}[t]

\fbox{
\begin{subfigure}{0.97\textwidth}
\centering
\begin{subfigure}{\MiniFigureWidth\textwidth}
  \centering
  \includegraphics[trim=35 0 35 0,clip,width=.99\linewidth ]{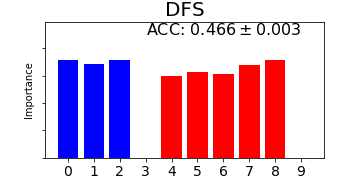}
\end{subfigure}
\begin{subfigure}{\MiniFigureWidth\textwidth}
  \centering
  \includegraphics[trim=35 0 35 0,clip,width=.99\linewidth ]{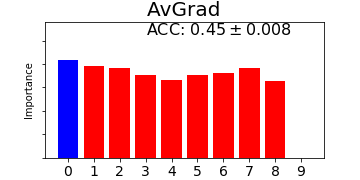}
\end{subfigure}
\begin{subfigure}{\MiniFigureWidth\textwidth}
  \centering
  \includegraphics[trim=35 0 35 0,clip,width=.99\linewidth ]{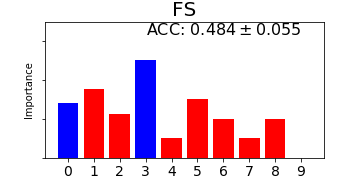}
\end{subfigure}
\begin{subfigure}{\MiniFigureWidth\textwidth}
  \centering
  \includegraphics[trim=35 0 35 0,clip,width=.99\linewidth ]{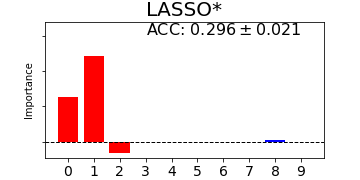}
\end{subfigure}
\begin{subfigure}{\MiniFigureWidth\textwidth}
  \centering
  \includegraphics[trim=35 0 35 0,clip,width=.99\linewidth ]{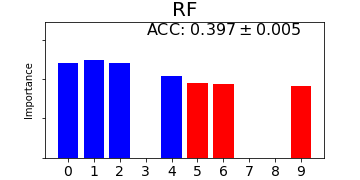}
\end{subfigure}
\begin{subfigure}{\MiniFigureWidth\textwidth}
  \centering
  \includegraphics[trim=35 0 35 0,clip,width=.99\linewidth ]{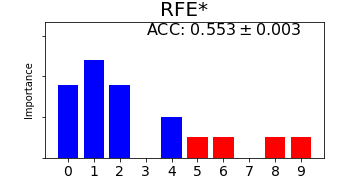}
\end{subfigure}
\begin{subfigure}{\MiniFigureWidth\textwidth}
  \centering
  \includegraphics[trim=35 0 35 0,clip,width=.99\linewidth ]{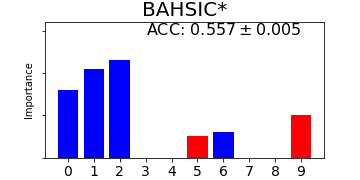}
\end{subfigure}
\begin{subfigure}{\MiniFigureWidth\textwidth}
  \centering
  \includegraphics[trim=35 0 35 0,clip,width=.99\linewidth ]{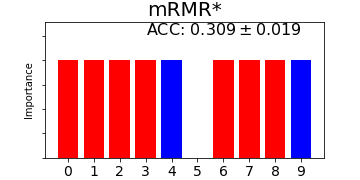}
\end{subfigure}
\begin{subfigure}{\MiniFigureWidth\textwidth}
  \centering
  \includegraphics[trim=35 0 35 0,clip,width=.99\linewidth ]{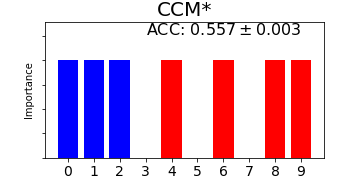}
\end{subfigure}
\begin{subfigure}{\MiniFigureWidth\textwidth}
  \centering
  \includegraphics[trim=35 0 35 0,clip,width=.99\linewidth ]{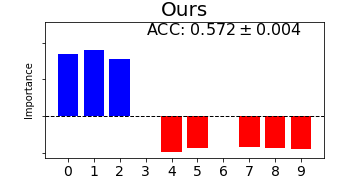}
\end{subfigure}
\caption{}
\end{subfigure}
}

\fbox{
\begin{subfigure}{0.97\textwidth}
\centering
\begin{subfigure}{\MiniFigureWidth\textwidth}
  \centering
  \includegraphics[trim=35 0 35 0,clip,width=.99\linewidth ]{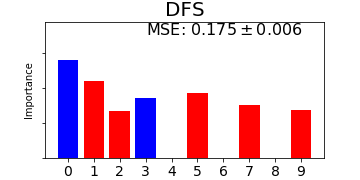}
\end{subfigure}
\begin{subfigure}{\MiniFigureWidth\textwidth}
  \centering
  \includegraphics[trim=35 0 35 0,clip,width=.99\linewidth ]{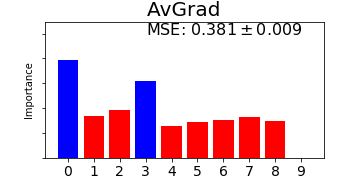}
\end{subfigure}
\begin{subfigure}{\MiniFigureWidth\textwidth}
  \centering
  \includegraphics[trim=35 0 35 0,clip,width=.99\linewidth ]{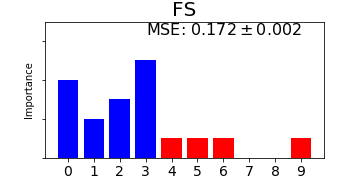}
\end{subfigure}
\begin{subfigure}{\MiniFigureWidth\textwidth}
  \centering
  \includegraphics[trim=35 0 35 0,clip,width=.99\linewidth ]{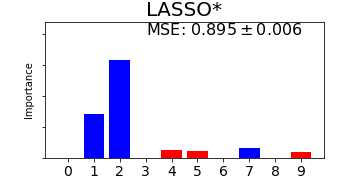}
\end{subfigure}
\begin{subfigure}{\MiniFigureWidth\textwidth}
  \centering
  \includegraphics[trim=35 0 35 0,clip,width=.99\linewidth ]{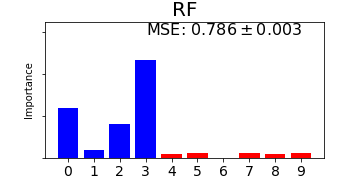}
\end{subfigure}
\begin{subfigure}{\MiniFigureWidth\textwidth}
  \centering
  \includegraphics[trim=35 0 35 0,clip,width=.99\linewidth ]{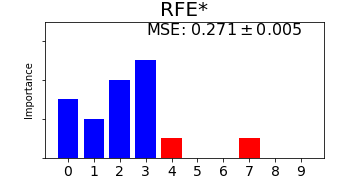}
\end{subfigure}
\begin{subfigure}{\MiniFigureWidth\textwidth}
  \centering
  \includegraphics[trim=35 0 35 0,clip,width=.99\linewidth ]{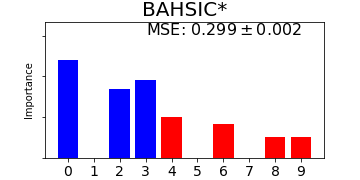}
\end{subfigure}
\begin{subfigure}{\MiniFigureWidth\textwidth}
  \centering
  \includegraphics[trim=35 0 35 0,clip,width=.99\linewidth ]{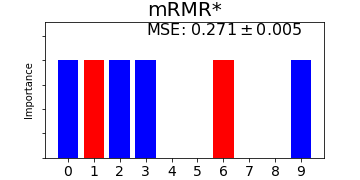}
\end{subfigure}
\begin{subfigure}{\MiniFigureWidth\textwidth}
  \centering
  \includegraphics[trim=35 0 35 0,clip,width=.99\linewidth ]{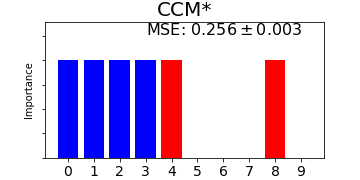}
\end{subfigure}
\begin{subfigure}{\MiniFigureWidth\textwidth}
  \centering
  \includegraphics[trim=35 0 35 0,clip,width=.99\linewidth ]{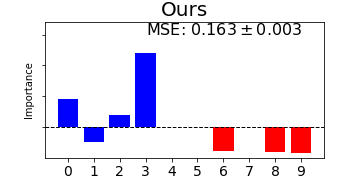}
\end{subfigure}
\caption{}
\end{subfigure}
}

\fbox{
\begin{subfigure}{0.97\textwidth}
\centering
\begin{subfigure}{\MiniFigureWidth\textwidth}
  \centering
  \includegraphics[trim=35 0 35 0,clip,width=.99\linewidth ]{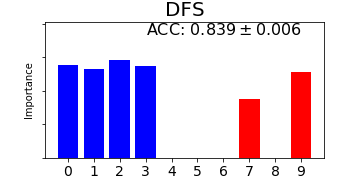}
\end{subfigure}
\begin{subfigure}{\MiniFigureWidth\textwidth}
  \centering
  \includegraphics[trim=35 0 35 0,clip,width=.99\linewidth ]{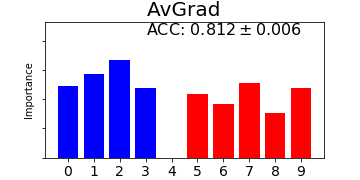}
\end{subfigure}
\begin{subfigure}{\MiniFigureWidth\textwidth}
  \centering
  \includegraphics[trim=35 0 35 0,clip,width=.99\linewidth ]{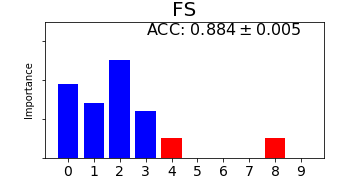}
\end{subfigure}
\begin{subfigure}{\MiniFigureWidth\textwidth}
  \centering
  \includegraphics[trim=35 0 35 0,clip,width=.99\linewidth ]{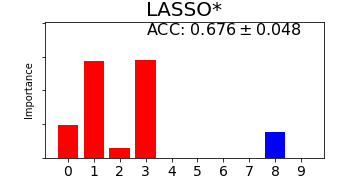}
\end{subfigure}
\begin{subfigure}{\MiniFigureWidth\textwidth}
  \centering
  \includegraphics[trim=35 0 35 0,clip,width=.99\linewidth ]{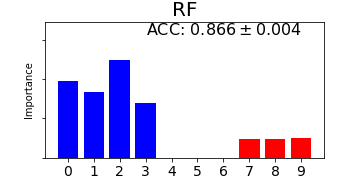}
\end{subfigure}
\begin{subfigure}{\MiniFigureWidth\textwidth}
  \centering
  \includegraphics[trim=35 0 35 0,clip,width=.99\linewidth ]{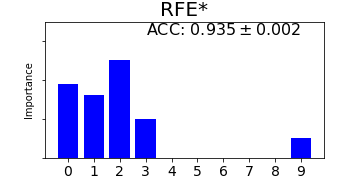}
\end{subfigure}
\begin{subfigure}{\MiniFigureWidth\textwidth}
  \centering
  \includegraphics[trim=35 0 35 0,clip,width=.99\linewidth ]{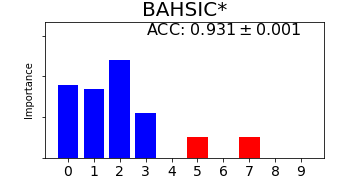}
\end{subfigure}
\begin{subfigure}{\MiniFigureWidth\textwidth}
  \centering
  \includegraphics[trim=35 0 35 0,clip,width=.99\linewidth ]{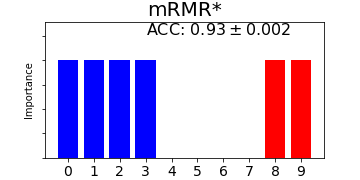}
\end{subfigure}
\begin{subfigure}{\MiniFigureWidth\textwidth}
  \centering
  \includegraphics[trim=35 0 35 0,clip,width=.99\linewidth ]{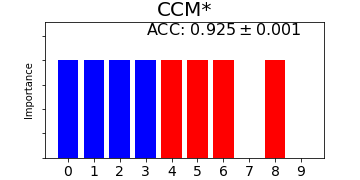}
\end{subfigure}
\begin{subfigure}{\MiniFigureWidth\textwidth}
  \centering
  \includegraphics[trim=35 0 35 0,clip,width=.99\linewidth ]{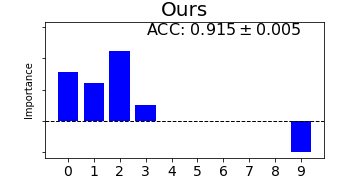}
\end{subfigure}
\caption{}
\end{subfigure}
}
\caption{5-fold cross-validation results (mean$\pm$str) on synthetic datasets ($s=5, d=10$). (a) XOR 4-way classification. (b) Nonlinear regression. (c) Binary classification. * refers to a filtering method, and blue/red colors indicate a feature selected in all 5 folds/fewer folds, respectively.}
\label{fig:synthetic}
\end{figure}

Fig. \ref{fig:synthetic} shows the feature selection and FIR results yielded by different methods regarding top 5 features on 3 synthetic datasets where the FIR scores are normalized in each method and the equal FIR score is set to all the features selected by those methods without considering FIR. It is observed from Fig. \ref{fig:synthetic} that our approach always finds out those relevant features in all 5 folds and does FIR properly by assigning negative scores (gradients), meaning unimportant, to irrelevant features. For the 4-way classification, DFS, RF, RFE, BAHSIC and CCM also find 3 relevant features in all 5 folds but others fail as shown in Fig. \ref{fig:synthetic}(a) although mRMR and CCM cannot yield FIR scores. In terms of accuracy, ours outperforms all other 9 methods despite the fact that DFS, AvGrad and RF work directly on the full feature set. For the nonlinear regression, FS, RF, RFE and CCM also select 4 relevant features in all 5 folds but ours yields the least MSE as shown in Fig. \ref{fig:synthetic}(b). For the binary classification, all the methods apart from LASSO find 4 relevant features in all 5 folds, as shown in Fig. \ref{fig:synthetic}(c). For this dataset, those state-of-the-art filtering methods yield better accuracy than others and the accuracy resulting from ours is slightly worse but comparable to those. In terms of FIR on all relevant features, ours is entirely consistent with those yielded by RF but performs significantly better than RF on 3 datasets. In comparison to the existing FIR methods for deep learning, ours always outperforms DFS, AvGrad and FS on 3 datasets in terms of both FIR and learning performance.

\subsection{Benchmark Data}
\label{sect:benchmark}

We further evaluate our approach on several well-known benchmark datasets from two different perspectives; i.e., explainability of FIR and learning performance on supervised feature selection. Evaluation on more benchmark datasets can be seen from Sect. B in Supplementary Materials.

\begin{figure}[t]
\centering
\begin{subfigure}{\MnistFigureWidth\textwidth}
  \centering
  \includegraphics[width=1.0\linewidth]{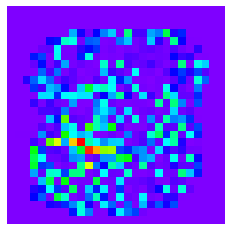}
  	\caption{}
\end{subfigure}
\begin{subfigure}{\MnistFigureWidth\textwidth}
  \centering
  \includegraphics[width=1.0\linewidth]{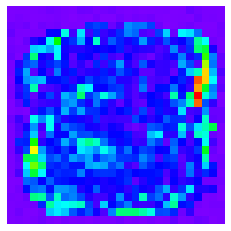}
  	\caption{}
\end{subfigure}
\begin{subfigure}{\MnistFigureWidth\textwidth}
  \centering
  \includegraphics[width=1.0\linewidth]{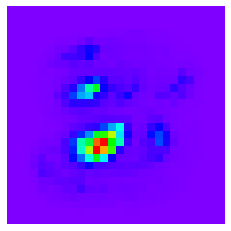}
  	\caption{}
\end{subfigure}
	\begin{subfigure}{\MnistFigureWidth\textwidth}
	  \centering
	  \includegraphics[width=1.0\linewidth]{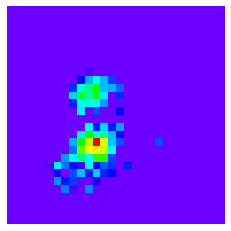}
	  	  \caption{}
	\end{subfigure}
	\begin{subfigure}{\MnistFigureWidth\textwidth}
	  \centering
	  \includegraphics[width=1.0\linewidth]{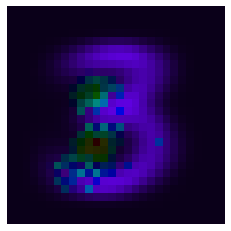}
	  \caption{}
	\end{subfigure}
	\begin{subfigure}{\MnistFigureWidth\textwidth}
	  \centering
	  \includegraphics[width=1.0\linewidth]{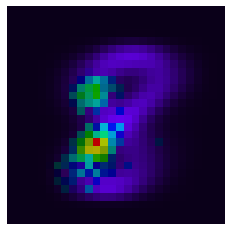}
	  	  \caption{}
	\end{subfigure}
\begin{subfigure}{0.9\textwidth}
  \centering
  \includegraphics[width=1.0\linewidth]{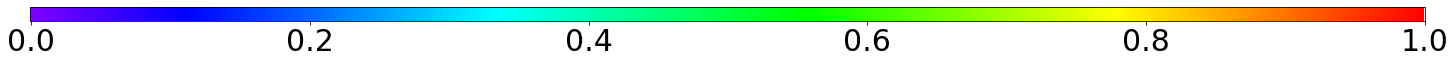}
\end{subfigure}
\caption{Feature importance maps yielded by different FIR methods. (a) DFS. (b) AvGrad. (c) RF. (d-f) Ours and our map superimposed on the mean images of ``3'' and ``8'', respectively, for clarity.}
\label{fig:mnist_importances}
\end{figure}

\noindent
\textbf{MNIST Dataset} \cite{Mnist}. To demonstrate the explainability of FIR via visual inspection, we employ an MNIST subset of hard-to-distinguish digits ``3'' and ``8'' for binary classification. The information on this subset is summarized in Table \ref{tab:benchmark}.
For comparison, we also apply 3 embedding methods, DFS, AvGrad and RF to this subset. To see the explainablity of FIR, we adopt the same full-connected MLP instead of CNN in DFS, AvGrad and the operator net in ours ($s=85, d=784$). The setting ensures that no other mechanisms like convolution/pooling layers can help a model automatically extract salient features for FIR.
As a result, the accuracies yielded by DFS, AvGrad, RF and \textbf{ours} on the test data are  $97.42\pm 0.30\%$, $99.27\pm 0.04\%$, $98.84\pm 0.03 \%$ and $\bf{99.31\pm 0.08\%}$, respectively, where ours and DFS use 85 and 212 features, respectively, but AvGrad and RF need all 784 features. For visual inspection, we normalize the FIR scores achieved by different methods to the same range and illustrate typical feature importance maps produced by 4 methods in a fold in Fig. \ref{fig:mnist_importances}.  It is observed from Fig. \ref{fig:mnist_importances}(a),(b) that DFS and AvGrad, two FIR methods for deep learning, do not produce explainable maps. In contrast, it is evident from Fig. \ref{fig:mnist_importances}(d-f) that ours yields a meaningful map where those features (pixels) that distinguish between ``3'' and ``8'' images are vividly highlighted in terms of their importance. Again, ours yields a map similar to that of RF (c.f. Fig. \ref{fig:mnist_importances}(c))  but outperforms this off-the-shelf FIR method.

\begin{table}[t]
\begin{center}
\caption{Information on benchmark and real-world datasets used in our experiments.}
\begin{tabular}{|c|c|c|c|c|c|c| }
\hline
 Data Set & MNIST & glass & vowel & TOX-171 & yale & Enhancer–Promoter \\
\hline
\hline
\#Features & 784 & 10 & 10 & 5784 & 1024 ($32 \times 32$) & 102\\
\#Classes & 2 & 6 & 11 & 4 &15 & 3\\
\#Training & 11,982 & 150  & 742 & 137 & 132 & 5,756\\
\#Testing     & 1,984 & 64  & 248 & 34 & 33 & 2,878 \\
\hline
\end{tabular}
\label{tab:benchmark}
\end{center}
\end{table}

\begin{figure}[h]
\centering

	\begin{subfigure}{.245\textwidth}
	  \centering
  \includegraphics[width=.99\linewidth]{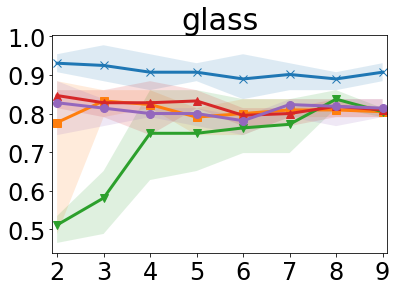}
	\end{subfigure}
	\begin{subfigure}{.245\textwidth}
	  \centering
  \includegraphics[width=.99\linewidth]{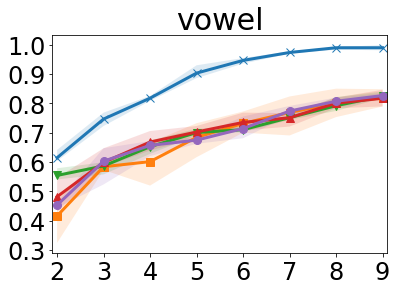}
	\end{subfigure}
\begin{subfigure}{.245\textwidth}
  \centering
  \includegraphics[width=.99\linewidth]{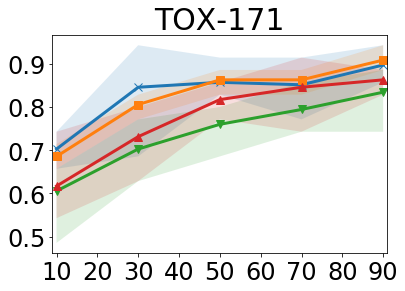}
\end{subfigure}
\begin{subfigure}{.245\textwidth}
  \centering
  \includegraphics[width=.99\linewidth]{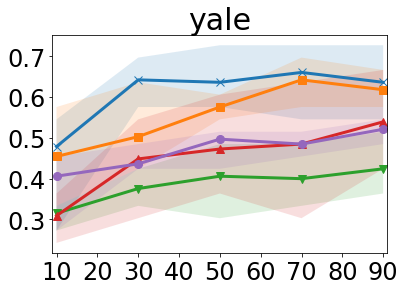}
\end{subfigure}

\begin{subfigure}{0.5\textwidth}
  \centering
 \includegraphics[width=.99\linewidth]{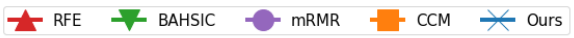}
\end{subfigure}
\caption{Classification accuracies (vertical axis) yielded by the supervised feature selection methods and ours for different numbers of selected features
(horizontal axis) on 4 benchmark datasets.}
\label{fig:benchmark_res}
\end{figure}

\noindent
\textbf{Feature Selection Benchmark}. We further conduct the evaluation in feature selection. As our approach has the same setting as used in the supervised feature selection methods, we compare ours to those strong supervised feature selection methods, RFE, BAHSIC, mRMR and CCM, on four benchmark datasets: \texttt{glass} \cite{Dua2019UCI}, \texttt{vowel} \cite{Dua2019UCI}, \texttt{TOX-171} \cite{li2018featureselectdata} and \texttt{yale} \cite{UCSD2001yale}, as summarized in Table \ref{tab:benchmark}.
For our model, we employ MLPs to implement the operator for \texttt{glass}, \texttt{vowel} and \texttt{TOX-171} but a CNN to carry out the operator for \texttt{yale} to demonstrate the flexibility of our dual-net architecture. By following the setting used in \cite{chen2017CCM}, we employ kernel SVMs for classification on features selected by 4 filtering methods.
It is evident from Fig. \ref{fig:benchmark_res} that ours substantially outperforms all others on \texttt{glass}, \texttt{vowel} and \texttt{yale} with a large margin. Overall, ours yields results comparable to the strongest performer, CCM, on \texttt{TOX-171} where there are 5,700+ features but only 109 training examples for parameter estimate in each of 5 folds, which is very challenging for deep learning.

\subsection{Real-world Data}
\label{sect:realdata}

\begin{figure}[t]
\centering
	\begin{subfigure}{1.0\textwidth}
	  \centering
  \includegraphics[width=\linewidth]{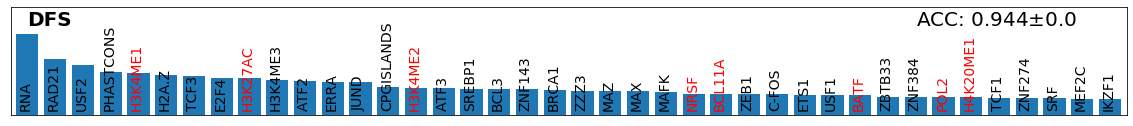}
	\end{subfigure}
		\begin{subfigure}{1.0\textwidth}
	  \centering
  \includegraphics[width=\linewidth]{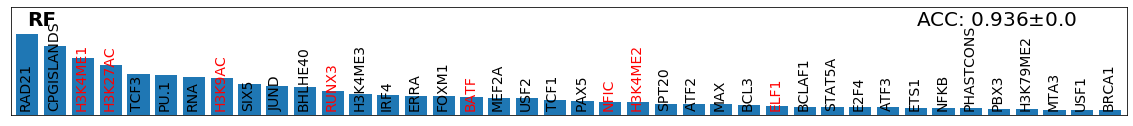}
	\end{subfigure}
		\begin{subfigure}{1.0\textwidth}
	  \centering
  \includegraphics[width=\linewidth]{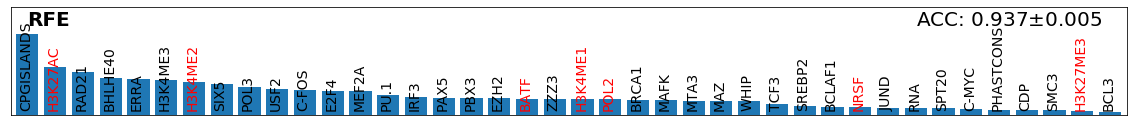}
	\end{subfigure}
	\begin{subfigure}{1.0\textwidth}
	  \centering
  \includegraphics[width=\linewidth]{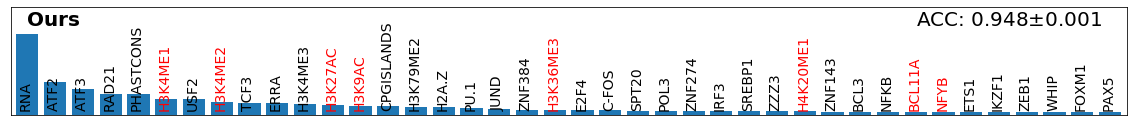}
	\end{subfigure}
	
\caption{Accuracy and FIR scores yielded by different methods on GM12878 Cell line (200 bp).}
\label{fig:enhance-promoter}
\end{figure}

We finally evaluate our approach on a real-world enhancer–promoter data, a challenging task that classifies the function of DNA sequences into enhancer, promoter and background \cite{anderson2014enhancer}. As listed in Table \ref{tab:benchmark}, the data used in this experiment are sampled from annotated DNA regions of GM12878 cell line (200 dp), the same as used in DFS \cite{Li2016DFS}, a feature selection method dedicated to this task. For comparison, we also apply DFS as well as RF and RFE, 2 strongest FIR methods manifested in our experiments, to this dataset. The same MLP architecture is used in implementing DFS and the operator net in our model.
Fig. \ref{fig:enhance-promoter} shows accuracies and the FIR scores of top $s=35$ out of $d=102$ features yielded by 4 different methods and those features colored in red correspond to the genes of which functions are well known in medicine and genetics literature (see Table 2 in \cite{Li2016DFS} for details). In terms of accuracy, ours is comparable to DFS and slightly better than RF and RFE, where the test accuracies of RFE and ours are based on $35$ features but the accuracies of RF and DFS are achieved with all 102 and 94 features selected by DFS via weight shrinkage, respectively. As seen in Fig. \ref{fig:enhance-promoter}, those top features ranked by DFS  and ours, two deep learning methods, appear quite similar but significantly different from those top features ranked by RF and RFE. The biological implication resulting from the results shown in Fig. \ref{fig:enhance-promoter} is worth investigating further from a biological/medical perspective.
More results on this dataset can be found from Sect. C in Supplementary Materials.

Regarding our alternate learning algorithm, our empirical studies suggest that it generally converges by reaching a local optimum (see Sects. B and C in Supplementary Materials for details).

\section{Discussion}

In general, our idea is motivated by RF \cite{breiman2001RF} and the dropout regularization \cite{srivastava2014dropout}; our exploration-exploitation strategy (c.f. Sect. \ref{subsect:learn}) allows for the simultaneous use of different feature subsets and dropout of input ``nodes'' randomly during learning. As the joint use of multiple feature subsets in learning leads to more training examples of fewer features randomly, our approach could provide an alternative way to improve the generalization in deep learning when the limited training examples are available even though FIR/feature selection is not of interest in such application scenarios.

Also, we want to make a connection between our proposed approach and evolutionary computation in terms of feature selection \cite{xue2016fectureselectionec}. In our approach, a single deep learning model, operator, works on different feature masks simultaneously during learning to carry out the functionality of a population of individual learning models in evolutionary computation. Instead of purely stochastic operations, mutation and crossover, on individual learners in a population used in evolutionary computation, our selector carried out by another single deep learning model uses a more efficient gradient-guided local stochastic search strategy to reduce the search space for combinatorial optimization. In general, our approach bears the spirit of evolution computation but addresses the combinatorial optimization issue in an entirely distinct manner, which leads to a more effective yet efficient approach to populationwise FIR and feature selection.

Our proposed approach is scalable to big data and easily makes use of any state-of-the-art deep learning techniques to be our component models for populationwise FIR and feature selection. In terms of computational complexity, however, our approach suffers from a high computational burden in training due to use of the dual-net architecture involving two deep learning models and the alternate learning procedure (see Sect. C in Supplementary Materials for details). Nevertheless, the computational load issue in our approach could be addressed (at least alleviated) by the latest development in deep learning, e.g., EfficientNet \cite{tan2019efficientnet}.

Our approach can be applied to the generic populationwise feature selection problem that needs to find out an optimal feature subset from $\sum_{s=1}^{d-1}\binom{d}{s}$ subsets for a feature set of $d$ features. Instead of a direct search of the entire subset space, we adopt a strategy that makes our model work in parallel on different subset sizes, the same as used in the state-of-the-art supervised feature selection methods, e.g., CCM \cite{chen2017CCM}. To this end, however, our approach might have a higher computational burden than those kernel-based methods in learning.
Also, our approach is extensible to group-based FIR and feature selection by introducing the group feature constraints to our stochastic local search procedure (c.f. Sect. \ref{subsect:learn}), which would overcome the limitation of linear models, e.g., group LARS/LASSO \cite{yuan2006groupLASSO}, in capturing the complex functional dependency between group input features and targets. Furthermore, our proposed dual-net architecture can also be extended to unsupervised feature selection by carrying out the operator with an autoencoder-like learning model.

In conclusion, we propose a dual-net neural architecture along with an alternate learning algorithm to enable deep learning to work effectively for FIR and feature selection. A thorough evaluation manifests that our approach outperforms several state-of-the-art FIR and supervised feature selection methods. In our ongoing work, we would extend our approach to instancewise FIR,  group and unsupervised feature selection scenarios and explore its potential in challenging real applications.

\section*{Broader Impact}

This research does not involve any issues directly regarding ethical aspects and future societal consequences. In the future, our approach presented in this paper might be applied in different domains, e.g., medicine and life science, where ethical aspects and societal consequences might have to be considered.

\section*{Acknowledgement}

The authors are grateful to three anonymous reviewers for their comments. In particular, the authors would thank the anonymous reviewer who offers us an insight by understanding our contributions from an evolutionary computation perspective.

\small
\bibliographystyle{unsrt}
\bibliography{mybib}

\begin{thebibliography}{10}

\bibitem{samek2017explainable}
W.~Samek, T.~Wiegand, and K.~R. M{\"u}ller.
\newblock Explainable artificial intelligence: Understanding, visualizing and
  interpreting deep learning models.
\newblock {\em arXiv preprint arXiv:1708.08296}, 2017.

\bibitem{guyon2002RFE}
I.~Guyon, J.~Weston, and S.~Barnhill.
\newblock Gene selection for cancer classification using support vector
  machines.
\newblock {\em Machine Learning}, 46(2):389--422, 2002.

\bibitem{guyon2003SFintro}
I.~Guyon and A.~Elisseeff.
\newblock An introduction to variable and feature selection.
\newblock {\em Journal of Machine Learning Research}, 3(3):1157--1182, 2003.

\bibitem{song2007BAHSIC1}
L.~Song, A.~Smola, A.~Gretton, K.~M. Borgwardt, and J.~Bedo.
\newblock Supervised feature selection via dependence estimation.
\newblock In {\em International Conference on Machine learning}, pages
  823--830, 2007.

\bibitem{song2012BAHSIC2}
L.~Song, A.~Smola, A.~Gretton, J.~Bedo, and K.~Borgwardt.
\newblock Feature selection via dependence maximization.
\newblock {\em Journal of Machine Learning Research}, 13(5):1393--1434, 2012.

\bibitem{hooker2019benchmark}
S.~Hooker, D.~Erhan, P.~J. Kindermans, and B.~Kim.
\newblock A benchmark for interpretability methods in deep neural networks.
\newblock In {\em Advances in Neural Information Processing Systems}, 2019.

\bibitem{Li2016DFS}
Y.~Li, C.~Y. Chen, and W.~Wasserman.
\newblock Deep feature selection: Theory and application to identify enhancers
  and promoters.
\newblock {\em Journal of Computational Biology}, 23(5):322--326, 2016.

\bibitem{weston2003NP-hard-FS}
J.~Weston, A.~Elisseeff, and B.~Scholkopf.
\newblock Use of zero-norm with linear models and kernel methods.
\newblock {\em Journal of Machine Learning Research}, 3(5):1439--1461, 2003.

\bibitem{tibshirani1996LASSO}
R.~Tibshirani.
\newblock Regression shrinkage and selection via the {LASSO}.
\newblock {\em J. R. Stat. Soc. Series B Stat. Methodol.}, 58(2):267--288,
  1996.

\bibitem{zou2005Elasticnet}
H.~Zou and T.~Hastie.
\newblock Regularization and variable selection via the elastic net.
\newblock {\em J. R. Stat. Soc. Series B Stat. Methodol.}, 67(2):301--320,
  2005.

\bibitem{friedman2001els}
J.~Friedman, T.~Hastie, and R.~Tibshirani.
\newblock {\em The Element of Statistical Learning}.
\newblock Springer, Berlin, 2001.

\bibitem{hechtlinger2016Inputgradient}
H.~Yotam.
\newblock Interpretation of prediction models using the input gradient.
\newblock {\em ArXiv: 1611.07634}, 2016.

\bibitem{ribeiro2016should}
M.~T. Ribeiro, S.~Singh, and C.~Guestrin.
\newblock Why should {I} trust you?: Explaining the predictions of any
  classifier.
\newblock In {\em Proceedings of the 22nd ACM SIGKDD international conference
  on knowledge discovery and data mining}, pages 1135--1144. ACM, 2016.

\bibitem{linden2019explanationbb}
I.~{van der Linden}, H.~Haned, and E.~Kanoulas.
\newblock Global aggregations of local explanations for black box models.
\newblock In {\em Proceedings of SIGIR Workshop on Fairness, Accountability,
  Confidentiality, Transparency, and Safety}, 2019.

\bibitem{skrlj2020fiesan}
B.~Skrlj, S.~Dzeroski, N.~Lavrac, and M.~Petkovic.
\newblock Feature importance estimation with self-attention networks.
\newblock In {\em Proceedings of 24th European Conference on Artificial
  Intelligence}, 2020.

\bibitem{breiman2001RF}
L.~Breiman.
\newblock Random forests.
\newblock {\em Machine Learning}, 45(1):5--32, 2001.

\bibitem{chen2017CCM}
J.~Chen, M.~Stern, M.~Wainwright, and M.~Jordan.
\newblock Kernel feature selection via conditional covariance minimization.
\newblock In {\em Advances in Neural Information Processing Systems}, 2017.

\bibitem{peng2005mRMR}
H.~Peng, F.~Long, and C.~Ding.
\newblock Feature selection based on mutual information criteria of
  max-dependency, max-relevance, and min-redundancy.
\newblock {\em IEEE Transactions on Pattern Analysis and Machine Intelligence},
  27(8):1226--1238, 2005.

\bibitem{wolpert1997NFL}
D.~H. Wolpert and W.~G. Macready.
\newblock No free lunch theorems for optimization.
\newblock {\em IEEE Transactions on Evolutionary Computation}, 1(1):67--82,
  1997.

\bibitem{mengshoel2008noise}
O.~J. Mengshoel.
\newblock Understanding the role of noise in stochastic local search: Analysis
  and experiments.
\newblock {\em Artificial Intelligence}, 172(2):955--990, 2008.

\bibitem{Mnist}
Y.~LeCun, C.~Cortez, and C.~Burges.
\newblock {The {MNIST} Handwritten Digit Database}.
\newblock \url{http://yann.lecun.com/exdb/mnist/}.

\bibitem{Dua2019UCI}
D.~Dua and C.~Graff.
\newblock {UCI Machine Learning Repository}.
\newblock \url{http://archive.ics.uci.edu/ml}.

\bibitem{li2018featureselectdata}
J.~Li, K.~Cheng, S.~Wang, F.~Morstatter, R.~P. Trevino, J.~Tang, and H.~Liu.
\newblock Feature selection: A data perspective.
\newblock {\em ACM Computing Surveys}, 50(6):94, 2018.

\bibitem{UCSD2001yale}
{Yale Face Database}.
\newblock \url{http://vision.ucsd.edu/content/yale-face-database}.

\bibitem{anderson2014enhancer}
R.~Anderson, C.~Gebard, I.~Miguel-Escalada, et~al.
\newblock An atlas of active enhancers across human cell types and tissues.
\newblock {\em Nature}, 507:455--461, 2014.

\bibitem{srivastava2014dropout}
N.~Srivastava, G.~Hinton, A.~Krizhevsky, I.~Sutskever, and R.~Salakhutdinv.
\newblock Dropout: {A} simple way to prevent neural networks from overfitting.
\newblock {\em Journal of Machine Learning Research}, 15(8):1929--1958, 2014.

\bibitem{xue2016fectureselectionec}
B.~Xue, M.~Zhang, W.~N. Browne, and X.~Yao.
\newblock Survey on evolutionary computation approaches to feature selection.
\newblock {\em IEEE Transactions on Evolutionary Computation}, 20(4):606--626,
  2016.

\bibitem{tan2019efficientnet}
M.~Tan and Q.~V. Le.
\newblock {EfficientNet}: Rethinking model scaling for convolutional neural
  networks.
\newblock In {\em Proceedings of the 36th International Conference on Machine
  Learning}, 2019.

\bibitem{yuan2006groupLASSO}
M.~Yuan and Y.~Lin.
\newblock Model selection and estimation in regression with grouped variables.
\newblock {\em J. R. Stat. Soc. Series B Stat. Methodol.}, 68(1):49--67, 2006.

\end{thebibliography}


\begin{thebibliography}{10}

\bibitem{kingma2014Adam}
D.~P. Kingma and J.~Ba.
\newblock Adam: A method for stochastic optimization.
\newblock In {\em Proceedings of International Conference for Learning
  Representations}, pages 585--592, 2015.

\bibitem{tensorflow2015-whitepaper}
Mart\'{\i}n Abadi, Ashish Agarwal, Paul Barham, Eugene Brevdo, Zhifeng Chen,
  Craig Citro, Greg~S. Corrado, Andy Davis, Jeffrey Dean, Matthieu Devin,
  Sanjay Ghemawat, Ian Goodfellow, Andrew Harp, Geoffrey Irving, Michael Isard,
  Yangqing Jia, Rafal Jozefowicz, Lukasz Kaiser, Manjunath Kudlur, Josh
  Levenberg, Dan Man\'{e}, Rajat Monga, Sherry Moore, Derek Murray, Chris Olah,
  Mike Schuster, Jonathon Shlens, Benoit Steiner, Ilya Sutskever, Kunal Talwar,
  Paul Tucker, Vincent Vanhoucke, Vijay Vasudevan, Fernanda Vi\'{e}gas, Oriol
  Vinyals, Pete Warden, Martin Wattenberg, Martin Wicke, Yuan Yu, and Xiaoqiang
  Zheng.
\newblock {TensorFlow}: Large-scale machine learning on heterogeneous systems,
  2015.
\newblock Software available from tensorflow.org.

\bibitem{chollet2015keras}
Fran\c{c}ois Chollet et~al.
\newblock Keras.
\newblock \url{https://keras.io}, 2015.

\bibitem{scikit-learn}
F.~Pedregosa, G.~Varoquaux, A.~Gramfort, V.~Michel, B.~Thirion, O.~Grisel,
  M.~Blondel, P.~Prettenhofer, R.~Weiss, V.~Dubourg, J.~Vanderplas, A.~Passos,
  D.~Cournapeau, M.~Brucher, M.~Perrot, and E.~Duchesnay.
\newblock Scikit-learn: Machine learning in {P}ython.
\newblock {\em Journal of Machine Learning Research}, 12:2825--2830, 2011.

\bibitem{Li2016DFS}
Y.~Li, C.~Y. Chen, and W.~Wasserman.
\newblock Deep feature selection: Theory and application to identify enhancers
  and promoters.
\newblock {\em Journal of Computational Biology}, 23(5):322--326, 2016.

\bibitem{hechtlinger2016Inputgradient}
H.~Yotam.
\newblock Interpretation of prediction models using the input gradient.
\newblock {\em ArXiv: 1611.07634}, 2016.

\bibitem{guyon2003SFintro}
I.~Guyon and A.~Elisseeff.
\newblock An introduction to variable and feature selection.
\newblock {\em Journal of Machine Learning Research}, 3(3):1157--1182, 2003.

\bibitem{tibshirani1996LASSO}
R.~Tibshirani.
\newblock Regression shrinkage and selection via the {LASSO}.
\newblock {\em J. R. Stat. Soc. Series B Stat. Methodol.}, 58(2):267--288,
  1996.

\bibitem{breiman2001RF}
L.~Breiman.
\newblock Random forests.
\newblock {\em Machine Learning}, 45(1):5--32, 2001.

\bibitem{guyon2002RFE}
I.~Guyon, J.~Weston, and S.~Barnhill.
\newblock Gene selection for cancer classification using support vector
  machines.
\newblock {\em Machine Learning}, 46(2):389--422, 2002.

\bibitem{song2007BAHSIC1}
L.~Song, A.~Smola, A.~Gretton, K.~M. Borgwardt, and J.~Bedo.
\newblock Supervised feature selection via dependence estimation.
\newblock In {\em International Conference on Machine learning}, pages
  823--830, 2007.

\bibitem{song2012BAHSIC2}
L.~Song, A.~Smola, A.~Gretton, J.~Bedo, and K.~Borgwardt.
\newblock Feature selection via dependence maximization.
\newblock {\em Journal of Machine Learning Research}, 13(5):1393--1434, 2012.

\bibitem{peng2005mRMR}
H.~Peng, F.~Long, and C.~Ding.
\newblock Feature selection based on mutual information criteria of
  max-dependency, max-relevance, and min-redundancy.
\newblock {\em IEEE Transactions on Pattern Analysis and Machine Intelligence},
  27(8):1226--1238, 2005.

\bibitem{chen2017CCM}
J.~Chen, M.~Stern, M.~Wainwright, and M.~Jordan.
\newblock Kernel feature selection via conditional covariance minimization.
\newblock In {\em Advances in Neural Information Processing Systems}, 2017.

\bibitem{anderson2014enhancer}
R.~Anderson, C.~Gebard, I.~Miguel-Escalada, et~al.
\newblock An atlas of active enhancers across human cell types and tissues.
\newblock {\em Nature}, 507:455--461, 2014.

\bibitem{linden2019explanationbb}
I.~{van der Linden}, H.~Haned, and E.~Kanoulas.
\newblock Global aggregations of local explanations for black box models.
\newblock In {\em Proceedings of SIGIR Workshop on Fairness, Accountability,
  Confidentiality, Transparency, and Safety}, 2019.

\bibitem{skrlj2020fiesan}
B.~Skrlj, S.~Dzeroski, N.~Lavrac, and M.~Petkovic.
\newblock Feature importance estimation with self-attention networks.
\newblock In {\em Proceedings of 24th European Conference on Artificial
  Intelligence}, 2020.

\bibitem{ribeiro2016should}
M.~T. Ribeiro, S.~Singh, and C.~Guestrin.
\newblock Why should {I} trust you?: Explaining the predictions of any
  classifier.
\newblock In {\em Proceedings of the 22nd ACM SIGKDD international conference
  on knowledge discovery and data mining}, pages 1135--1144. ACM, 2016.

\bibitem{GARCIADIAZ20201916}
Pilar García-Díaz, Isabel Sánchez-Berriel, Juan~A. Martínez-Rojas, and
  Ana~M. Diez-Pascual.
\newblock Unsupervised feature selection algorithm for multiclass cancer
  classification of gene expression rna-seq data.
\newblock {\em Genomics}, 112(2):1916 -- 1925, 2020.

\bibitem{DING2005}
H.~Peng C.~Ding.
\newblock Minimum redundancy feature selection from microarray gene expression
  data.
\newblock {\em Journal of bioinformatics and computational biology}, 3(2):185
  -- 205, 2005.

\bibitem{MAHATA2007775}
Pritha Mahata and Kaushik Mahata.
\newblock Selecting differentially expressed genes using minimum probability of
  classification error.
\newblock {\em Journal of Biomedical Informatics}, 40(6):775 -- 786, 2007.
\newblock Intelligent Data Analysis in Biomedicine.

\bibitem{LIU20102763}
Huawen Liu, Lei Liu, and Huijie Zhang.
\newblock Ensemble gene selection for cancer classification.
\newblock {\em Pattern Recognition}, 43(8):2763 -- 2772, 2010.

\bibitem{BEST2015666}
Myron~G. Best, Nik Sol, Irsan Kooi, Jihane Tannous, Bart~A. Westerman,
  François Rustenburg, Pepijn Schellen, Heleen Verschueren, Edward Post, Jan
  Koster, Bauke Ylstra, Najim Ameziane, Josephine Dorsman, Egbert~F. Smit,
  Henk~M. Verheul, David~P. Noske, Jaap~C. Reijneveld, R.~Jonas~A. Nilsson,
  Bakhos~A. Tannous, Pieter Wesseling, and Thomas Wurdinger.
\newblock Rna-seq of tumor-educated platelets enables blood-based pan-cancer,
  multiclass, and molecular pathway cancer diagnostics.
\newblock {\em Cancer Cell}, 28(5):666 -- 676, 2015.

\bibitem{PIAO201739}
Yongjun Piao, Minghao Piao, and Keun~Ho Ryu.
\newblock Multiclass cancer classification using a feature subset-based
  ensemble from microrna expression profiles.
\newblock {\em Computers in Biology and Medicine}, 80:39 -- 44, 2017.

\bibitem{SAYGILI2018}
Ahmet Saygili.
\newblock Classification and diagnostic prediction of breast cancers via
  different classifiers.
\newblock 2:48--56, 12 2018.

\end{thebibliography}

\end{document}


\setlength{\fboxsep}{0pt}%
  \setlength{\fboxrule}{0pt}%

\maketitle

In this document, we present our experimental setup, more experimental results and the implementation of our alternate learning algorithm. Section \ref{sect:setup} describes the the detailed experimental setup for all the method including ours in our comparative study. Sections \ref{sect:benchmark} and \ref{sect:enhance-promoter}  report the typical learning behavior of our alternative learning algorithm and more experimental results . Section \ref{sect:pseudo} presents the pseudo code of our alternate learning algorithm.

\section{Experimental Setup}
\label{sect:setup}

In this section, we describe the details of experimental settings used in our experiments. In our experiments, we always use the \emph{grid search} along with \emph{5-fold cross-validation} on a training set to find out optimal hyperparameters involved in different learning methods. Below, we first present the detailed setup in our approach, then describe all the technical details of other learning methods used in our comparative studies on different datasets.

\subsection{Setup in Our Approach}

\begin{figure}[t]
\centering
  \includegraphics[trim=20 670 370 30, clip,width=0.5\linewidth]{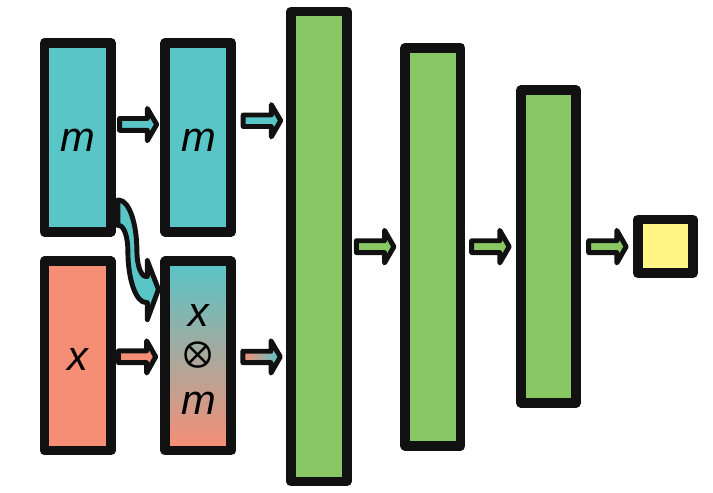}
  	\caption{The actual implementation of the operator net in our experiments to overcome the ``zero-value'' representation issue. As a result, both the selected features, $\pmb x \otimes \pmb m$, and the mask, $\pmb m$, used to indicate those selected features are concatenated as the input to the first hidden layer.}
\label{fig:example_ON}
\end{figure}

In our implementation of the operator net, we have to consider an issue concerning the differentiation between \emph{a selected feature of which value is zero} and \emph{any removed features masked with zero} due to the use of the binary masks in our work. Thus, we design an operator net architecture shown in Figure \ref{fig:example_ON}. Instead of feeding only the selected features, $\pmb x \otimes \pmb m$, to the first hidden layer, we concatenate the mask, $\pmb m$, used to indicate the selected features, and the selected features themselves, $\pmb x \otimes \pmb m$, to form the input feeded to the first hidden layer as illustrated in Figure \ref{fig:example_ON}. Thus, the dimension of the input to the first hidden layer is $2d$ rather than $d$ features described in the main text. It is worth mentioning that we had investigated other manners to tackle the aforementioned ``zero-value" issue, e.g., stipulating a value beyond the range of any features for a removed feature in $\pmb x \otimes \pmb m$. However, neither of those yields the better performance than the architecture presented above.

Moreover, there are specific settings in our approach due to technical reasons; e.g., the loss function used to train the selector net and the subtle technical details related to Phase II in our alternate learning algorithm, as described in Sect. 3.3 of the main text.

In our experiments, the loss function used to train the selector net presented in Eq.(2b) of the main text is actually replaced by a {\it weighted} loss as follows:
$$
{\cal L}_S ({\cal M}'; \varphi ) =  \frac 1 {2|{\cal M}'|} \sum_{\pmb m \in {\cal M}'}
w_{\pmb m}
\Big ( f_S(\varphi; \pmb m) - \frac 1 {|{\cal D}|} \sum_{({\pmb x}, {\pmb y}) \in {\cal D}}
l(\pmb x \otimes \pmb m, \pmb y; \theta) \Big )^2,
$$
where $w_{\pmb m} = 10, 5, 1$  when $\pmb m = {\pmb m}_{t,best}$ (the best performed subset found in the last step), $\pmb m = {\pmb m}_{t+1,opt}$ (the optimal subset generated in the current step), and $\pmb m$ is any of other subsets in ${\cal M}_{t+1}'$, respectively (c.f. Phase II-A in Sect. 3.3 of the main text). The above weighted selector loss exploits what has been learned so far in order to facilitate the stochastic local search in tackling the combinatorial optimization problem more effectively.

\begin{table}[t]
\caption{The optimal architectural hyperparameters of our dual-net learning model in our experiments. \textbf{ CNN$^*$}: there are
3 convolutional layers of 32, 64 and 128 channels and the corresponding kernel sizes are  5, 3, 3, respectively. Each of the convolutional layers is followed by a maximum pooling layer. Two input channels are used for coping with the ``zero-value" issue,  one
for the selected features, $\pmb x \otimes \pmb m$, and the other for the mask, $\pmb m$. (c.f. Figure \ref{fig:example_ON})
 Two dense layers of 30 neurons are top on the last convolutional layer. The output layer of 15 softmax neurons corresponds to 15 classes.}
\begin{center}
\makebox[\textwidth][c]{
\begin{tabular}{|c|c|c|}
    \hline
   {\bf Data Set} &  {\bf Operator Net} & {\bf Selector Net} \\
    \hline
     4-way Classification & $20 \rightarrow 60\rightarrow 30\rightarrow 20 \rightarrow  4$ & $10 \rightarrow 100\rightarrow 50\rightarrow 10 \rightarrow 1$  \\
     Nonlinear Regression & $20 \rightarrow 100\rightarrow 50\rightarrow 25 \rightarrow 1$ & $10 \rightarrow 100\rightarrow 50\rightarrow 10 \rightarrow 1$ \\
     Binary Classification & $20 \rightarrow 60\rightarrow 30\rightarrow 20 \rightarrow  1$ & $10 \rightarrow 100\rightarrow 50\rightarrow 10 \rightarrow 1$  \\
    \hline
     MNIST Subset & $1568 \rightarrow 500\rightarrow 250\rightarrow 100 \rightarrow 1$ & $784 \rightarrow 300\rightarrow 200\rightarrow 100 \rightarrow 1$  \\
     Glass &  $20 \rightarrow 50\rightarrow 25\rightarrow 10 \rightarrow 6$ &$10 \rightarrow 500\rightarrow 250\rightarrow 100 \rightarrow 1$ \\
     Vowel &  $20 \rightarrow 50\rightarrow 25\rightarrow 10 \rightarrow 11$ &$10 \rightarrow 500\rightarrow 250\rightarrow 100 \rightarrow 1$ \\
     TOX-171 &  $11568 \rightarrow 100\rightarrow 50\rightarrow 20\rightarrow 4$ & $5784 \rightarrow 500\rightarrow 250\rightarrow 100 \rightarrow 1$ \\
     Yale & \textbf{CNN$^*$} & $1024 \rightarrow 500\rightarrow 250\rightarrow 100 \rightarrow 1$ \\
     \hline
     Enhancer–Promoter & $204 \rightarrow 300\rightarrow 200\rightarrow 50 \rightarrow 3$ & $102 \rightarrow 500\rightarrow 250\rightarrow 100 \rightarrow 1$ \\
      \hline
     RNA-seq & $40528 \rightarrow 1000\rightarrow 500\rightarrow 200 \rightarrow 5$ & $20264 \rightarrow 500\rightarrow 250\rightarrow 100 \rightarrow 1$ \\
     \hline
\end{tabular}
}
\end{center}
\label{tab:architectures}
\end{table}

\begin{table}[t]
\caption{Other optimal hyperparameters of our dual-net learning model in our experiments.  $E_1$ is the number of SGD training batches (instead of epoches) in Phase I of the alternative learning. In Phase II.A,  $|{\cal M}_t'|$ refers to the number of different optimal subset candidates used in a single batch during the SGD learning. $f=\frac {|{\cal M}_{t,1}'|} {|{\cal M}_{t,2}'|}$ is the fraction that governs the exploitation-exploration trade-off in the selector learning, and $s_p$ indicates the number of elements perturbed.  For details of the alternative learning algorithm, see Section 3.3 in the main text.}

\begin{center}
\makebox[\textwidth][c]{
\begin{tabular}{|c|c|c|c|c|}
    \hline
   {\bf Data Set} &  $E_1$ & $|{\cal M}_t'|$ & $f$ & $s_p$ \\
    \hline
     4-way Classification & $6,000$ & $32$ & $0.5$ & $2$ \\
     Nonlinear Regression & $6,000$ & $32$ & $0.5$ & $2$ \\
     Binary Classification & $6,000$ & $32$ & $0.5$ & $2$ \\
    \hline
     MNIST Subset & $10,000$ & $32$ & $0.5$ & $5$ \\
     Glass & $10,000$ & $128$ & $0.5$ & $2$  \\
     Vowel & $6,000$ & $128$ & $0.5$ & $2$ \\
     TOX-171 & $1,500$   & $128$ & $0.5$ & $5$  \\
     Yale  & $4,500$  & $128$ & $0.5$ & $5$ \\
     \hline
     Enhancer–Promoter  & $10,000$ & $64$ & $0.5$ & $5$ \\
      \hline
     RNA-seq & $8000$ & $32$ & $0.5$ & $5$ \\
     \hline
\end{tabular}
}
\end{center}
\label{tab:hyperparameters}
\end{table}

In our experiments, the 3-step validation procedure used for \emph{generation of the optimal subset} would ensure its optimality within those feature subset candidates (c.f. Phrase II.A in Sect. 3.3 of the main text). However, the condition to exit from the loop of repeating steps i)-iii) may not be always satisfied. In our experiments, we hence set the maximum number of repetition in this test to \emph{5} iterations so that the subset optimality validation procedure always ends up to \emph{five} iterations. In addition, the parameter update of the operator and the selector nets in Phase II is  done in different frequencies in the alternate learning; i.e., the parameters in the operator net are updated once on each batch in Phase II.B, while the parameters in the selector net are updated once on every 8 batches in Phase II.A.

We employ MLPs (CNNs) of the sigmoid (ReLu) neurons to carry out the operator net and MLPs with the sigmoid neurons for the selector net in our dual-net architecture. For training MLPs (CNNs), we adopt the \emph{Adam optimizer} (Adam with Nestrov momentum for the operator net)\cite{kingma2014Adam}  via the stochastic gradient descent (SGD) procedure. Early stopping is used based on the losses evaluated on the validation data\footnote{In our alternate learning procedure, we use the operator loss incurred by the optimal subset, ${\pmb m}_{t,opt}$, on the validation set for early stopping. For clarity and details, see Section \ref{sect:behavior}.}.  All the optimal hyperparameters used in our experiments are summarized in Tables \ref{tab:architectures} and \ref{tab:hyperparameters}, respectively.

\subsection{Optimal Hyperparameters in Other Methods}

\begin{table}[t]
\caption{Optimal regularization hyperparameters, $\lambda$, used in LASSO on different datasets in our experiments.}

\begin{center}
\makebox[\textwidth][c]{
\begin{tabular}{|c|c|c|c|c|c|}
\hline
{\bf Data Set} &  Fold-1 & Fold-2 & Fold-3 &  Fold-4 & Fold-5 \\
\hline
4-way classification&0.055&0.056&0.046&0.008&0.042\\
Nonlinear Regression&0.001&0.0&0.053&0.001&0.0\\
Binary Classification&0.025&0.07&0.081&0.02&0.039\\
\hline
\end{tabular}
}
\end{center}
\label{tab:lasso_params}

\end{table}

\begin{table}[ht]
\caption{Optimal hyperparameters (\#tree, depth) for RF on different datasets in our experiments.}

\begin{center}
\makebox[\textwidth][c]{
\begin{tabular}{|c|c|c|c|c|c|}
\hline
 {\bf Data Set} &  Fold-1 & Fold-2 & Fold-3 &  Fold-4 & Fold-5 \\
\hline
4-way classification& (80, 14) & (70, 11) & (90, 15) & (150, 14) & (150, 12) \\
Nonlinear Regression& (50, 15) & (60, 15) & (50, 14) & (60, 10) & (50, 8) \\
Binary Classification& (60, 14) & (50, 10) & (90, 8) & (90, 15) & (160, 13) \\
\hline
MNIST& (200, 21) & (200, 20) & (210, 21) & (200, 21) & (200, 20)  \\
\hline
Enhancer-Promoter& (120, 11) & (120, 12) & (100, 15) & (150, 13) & (60, 14) \\
\hline
\end{tabular}
}
\end{center}
\label{tab:rf_params}
\end{table}

Other 9 methods have also been employed for a comparative study on different datasets. We strictly follow their original settings described in those papers. We implement deep learning algorithms with by ourselves with Tensorflow 2.0 \cite{tensorflow2015-whitepaper} and Keras \cite{chollet2015keras}. For other methods, we use the existing code in the Python scikit-Learn library \cite{scikit-learn} for FS, LASSO, RF, RFE or the authors' project website for BAHSIC\footnote{BAHSIC webpage: \url{https://www.cc.gatech.edu/~lsong/code.html}}, mRMR\footnote{PyMRMR library: \url{https://pypi.org/project/pymrmr/}} and CCM \footnote{CCM repository: \url{https://github.com/Jianbo-Lab/CCM}}. Our source code will be made available after the completion of review. Below, we summarize the actual optimal hyperparameters pertaining to those methods used in our comparative study.

\paragraph{Deep Feature Selection (DFS)} \cite{Li2016DFS}. For DFS, we use the MLPs of the architectures as same as that of the operator net in our dual-net architecture apart from the input layers for a given task, as shown in Table \ref{tab:architectures}. Instead of having the concatenation of the selected features and the mask indicating the selected features in our operator net, the DFS appends an additional one-to-one layer between the input and the first hidden layer.  Similarly, the sigmoid neurons are used in their modified MLPs and the Adam optimizer \cite{kingma2014Adam} is adopted for training MLPs via the SGD. The optimal regularization hyperparameter is $\lambda =0.01$ for 3 synthetic datasets and the MNIST subset after a grid search from a large range of $\lambda$. For the Enhance-Promoter dataset, the optimal hyperparameter is $\lambda = 0.008$. The rest of the parameters are kept the same as suggested in \cite{Li2016DFS}, which is $\lambda_2 = 1, \alpha_1 = 0.0001,  \alpha_2 = 0$.
Note that we implement the DFS code by ourselves with Tensorflow 2.0 and Keras since the authors' code is merely applicable to a specific dataset.

\paragraph{Average Input Gradient (AvGrad)} \cite{hechtlinger2016Inputgradient}. It is simply a post-processing method for feature importance ranking (FIR) based on a trained MLP, we employ the same MLP architecture as that of our operator net apart from the input and the Adam optimizer \cite{kingma2014Adam} via the SGD on 3 synthetic datasets and the MNIST subset, Table \ref{tab:architectures}.

\paragraph{Forward Selection (FS)} \cite{guyon2003SFintro}. For FS, we employ the MLPs as the base leaner in this wrapper method and the training procedure identical to those used in the AvGrad on 3 synthetic datasets. For FIR, FS always ranks the importance of an early selected feature higher than that of others selected later in the forward subset selection procedure.

\paragraph{LASSO} \cite{tibshirani1996LASSO}. We use the grid-search to find out the optimal regularization hyperparameter, $\lambda$, in LASSO.  The optimal hyperparameters found in 5 folds are listed in Table \ref{tab:lasso_params}.

\paragraph{Random Forest (RF)} \cite{breiman2001RF}. We use the grid-search to find out the optimal hyperparameters: number of decision trees and depth of the trees. We search from a range from 50 to 220 trees and between 7 and 24 in depth. The optimal hyperparameters found in 5 folds are listed in Table \ref{tab:rf_params}.

\paragraph{Recursive Feature Estimation (RFE)} \cite{guyon2002RFE}. We use 1 step for all the datasets apart from TOX-171 and Yale datasets where 5 steps are used. In our experiments, we adopt the default values for underlying estimators (linear SVM) with $C=1$ and $\gamma =  \frac{1}{n_{features} * \text{var}(X)}$.

\paragraph{Backward Elimination using HSIC (BAHSIC)} \cite{song2007BAHSIC1,song2012BAHSIC2}. A default hyperparemeter regarding the fraction of removed features in each iteration is set to 0.1 as suggested in their papers. In our experiments, we adopt the inverse kernels suggested in their papers and the BAHSIC webpage.

\paragraph{Minimal Redundancy Maximal Relevance Criterion (mRMR)} \cite{peng2005mRMR}. No hyperparemeter needs to be tuned in this method. In our experiments, we adopt he "MIQ" option suggested in the \texttt{PyMRMR} library.

\paragraph{Conditional Covariance Minimization (CCM)} \cite{chen2017CCM}. We use  $\epsilon=0.001$ for two synthetic classification datasets, 4-way and binary classification, and 4 benchmark datasets, Glass, Vowel, TOX-171 and Yale. For the nonlinear regression dataset, we use $\epsilon=0.1$. As all 7 datasets were used in the paper, we adopt the optimal hyperparameters reported in the paper and suggested in the CCM repository.

As CCM, RFE, BAHSIC, mRMR and LASSO are filtering methods for feature selection, we need to measure their performance based on another learning model. For CCM, RFE, BAHSIC and mRMR, we adopt the same setting used in \cite{chen2017CCM}, i.e., SVM/SVR with a Gaussian kernel of optimal hyperparameters: $C=1$ and $\gamma =  \frac{1}{n_{features} * \text{var}(X)}$. For LASSO, we use the same MLPs used in deep learning models, i.e., DFS, AvGrad and ours (operator net).

\section{More Results on Synesthetic and Benchmark Data}
\label{sect:benchmark}

In this section, we demonstrate the typical learning behavior of our alternate learning algorithm on different datasets, describe the detailed information of benchmark datasets used in our experiments and report more experimental results.

\begin{figure}[t]
\centering
\fbox {
\begin{subfigure}{\MnistFigureWidth\textwidth}
  \centering
  \includegraphics[width=\linewidth]{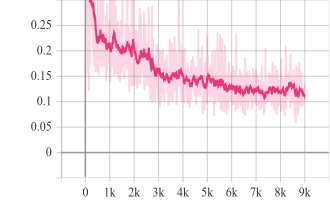}
  	\caption{selector-loss}
 \vspace*{4mm}
\end{subfigure}
}

\fbox {
	\begin{subfigure}{\MnistFigureWidth\textwidth}
	  \centering
  \includegraphics[width=.99\linewidth]{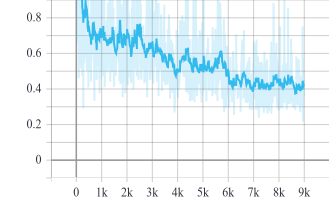}
	  	  \caption{operator-loss-train}
\vspace*{4mm}
	\end{subfigure}

	\begin{subfigure}{\MnistFigureWidth\textwidth}
	  \centering
  \includegraphics[width=.99\linewidth]{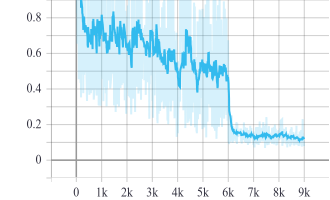}
	  	  \caption{operator-loss-train(${\pmb m}_{opt}$)}
\vspace*{4mm}
	\end{subfigure}

}
\fbox{
\begin{subfigure}{\MnistFigureWidth\textwidth}
	  \centering
  \includegraphics[width=.99\linewidth]{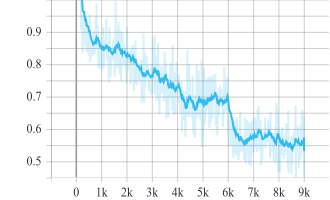}
	  	  \caption{operator-loss-val}
	\end{subfigure}

	\begin{subfigure}{\MnistFigureWidth\textwidth}
	  \centering
  \includegraphics[width=.99\linewidth]{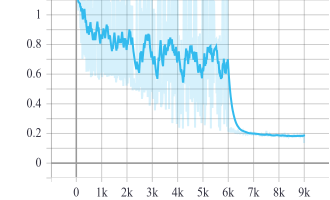}
	  	  \caption{operator-loss-val(${\pmb m}_{opt}$)}
	\end{subfigure}

}
\caption{\textbf{Synthetic nonlinear regression dataset}. Evolution of the operator and the selector losses in Phase II ($d=10, s=5$). The \texttt{x-axis} corresponds to the number of batches and \texttt{y-axis} refers to the loss statistics of 5 folds.
(a) The selector loss. (b) The operator loss on the training set. (c) The operator loss on the training set with ${\pmb m}_{opt}$ only.
 (d) The operator loss on the validation set.
(e) The operator loss on the validation set with ${\pmb m}_{opt}$ only.
Note that Phase II starts when the operator net has been trained for 6,000 batches in Phase I.}
\label{fig:synth_training}
\end{figure}

\subsection{Learning Behavior}
\label{sect:behavior}

As described in Section 3.3 of the main text, our alternate learning algorithm trains two learning models, operator and selector, simultaneously in an alternate manner; i.e., in Phase II, the learning behavior of the operator and the selector nets are mutually affected each other in each batch during the SGD learning. This is different from most of the existing deep learning algorithm that involves only a deep neural network to be trained. Therefore, we need to investigate how our proposed learning model behaves. Below, we exhibit the typical learning behavior in our alternative learning on 3 datasets, \emph{synthetic nonlinear regression}, \emph{MNIST subset} and \emph{Yale}.

Figure \ref{fig:synth_training} illustrates the learning behavior of the operator and the selector in terms of losses in Phase II on the synthetic \emph{nonlinear regression} dataset. It is observed from Figure \ref{fig:synth_training}(a) that the trained operator in Phase I provides informative training examples so that the averaged selector loss of 5 folds decreases monotonically as required. As evident in Figures \ref{fig:synth_training}(b) and \ref{fig:synth_training}(d), the averaged operator loss on training and validation sets further decreases steadily as the selector keeps offering more ``promising'' optimal mask candidates achieved by the stochastic local search for combinatorial optimization. It is clearly seen in Figures \ref{fig:synth_training}(b) and \ref{fig:synth_training}(d)  that at the beginning of Phase II (up to 1k batches), operator loss on both training and validation sets sharply decreases once the selector has been involved.  Also, the loss may be reduced substantially when an optimal mask is identified, as shown in Figure \ref{fig:synth_training}(d) (between 6k and 7k batches). Given the fact that at the end of Phase II.A for each iteration, we always achieve an optimal mask, ${\pmb m}_{t,opt}$. Thus, we can apply such optimal masks only to measure the operator loss. As a result, Figures  \ref{fig:synth_training}(c) and \ref{fig:synth_training}(e) illustrate the evolution of the operator loss evaluated with ${\pmb m}_{t,opt}$ only on training and validation sets. In contrast to the operator loss with all optimal mask (subset) candidates shown in  Figure  \ref{fig:synth_training}(d), the abrupt loss drop resulting from the identified optimal mask is much more visible in Figure  \ref{fig:synth_training}(e). Therefore, early stopping in our alternate learning algorithm is based on the operator loss evaluated with ${\pmb m}_{t,opt}$ only. Overall, Figure \ref{fig:synth_training} demonstrates that our alternate learning algorithm works well and eventually converges for this regression task.

\begin{figure}[ht]
\centering
\fbox {
\begin{subfigure}{\MnistFigureWidth\textwidth}
  \centering
  \includegraphics[width=\linewidth]{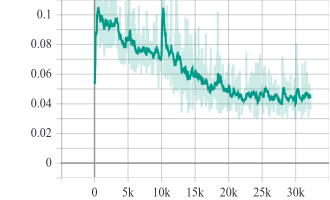}
  	\caption{selector-loss}
 \vspace*{4mm}
\end{subfigure}
}

\fbox{

	\begin{subfigure}{\MnistFigureWidth\textwidth}
	  \centering
  \includegraphics[width=.99\linewidth]{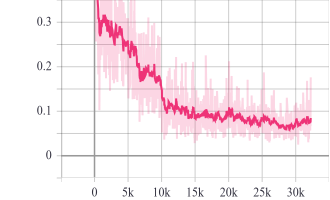}
	  	  \caption{operator-loss-train}
\vspace*{4mm}
	\end{subfigure}

	\begin{subfigure}{\MnistFigureWidth\textwidth}
	  \centering
  \includegraphics[width=.99\linewidth]{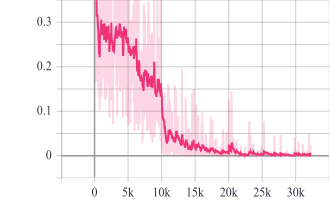}
	  	  \caption{operator-loss-train(${\pmb m}_{opt}$)}
\vspace*{4mm}
	\end{subfigure}

\begin{subfigure}{\MnistFigureWidth\textwidth}
	  \centering
  \includegraphics[width=.99\linewidth]{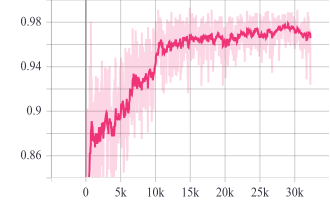}
	  	  \caption{operator-ACC-train}
\vspace*{4mm}
	\end{subfigure}
}

\fbox {
		\begin{subfigure}{\MnistFigureWidth\textwidth}
	  \centering
  \includegraphics[width=.99\linewidth]{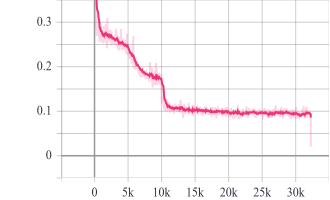}
	  	  \caption{operator-loss-val}
\vspace*{4mm}
	\end{subfigure}

	\begin{subfigure}{\MnistFigureWidth\textwidth}
	  \centering
  \includegraphics[width=.99\linewidth]{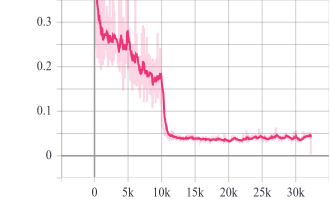}
	  	  \caption{operator-loss-val(${\pmb m}_{opt}$)}
\vspace*{4mm}
	\end{subfigure}
\begin{subfigure}{\MnistFigureWidth\textwidth}
	  \centering
  \includegraphics[width=.99\linewidth]{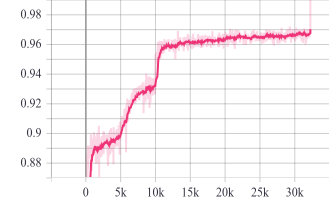}
	  	  \caption{operator-ACC-val}
\vspace*{4mm}
	\end{subfigure}
}
\caption{\textbf{MNIST Benchmark Subset}. Evolution of the operator and the selector losses in Phase II ($d=784, s=85$). The \texttt{x-axis} corresponds to the number of batches and \texttt{y-axis} refers to the loss statistics of 5 folds.
(a) The selector loss.
(b) The operator loss on the training set.
(c) The operator loss on the training set with ${\pmb m}_{opt}$ only.
(d) The classification accuracy evaluated on the training set.
(e) The operator loss on the validation set.
(f) The operator loss on the validation set with ${\pmb m}_{opt}$ only.
(g) The classification accuracy evaluated on the validation set.
Note that Phase II starts when the operator net has been trained for 10,000 batches in Phase I. The spike at the maximum batch in (e)-(g) correspond to the results evaluated on the test set with the trained operator net upon the completion of the alternate learning.}
\label{fig:training_mnist}
\end{figure}

\begin{figure}[ht]
\centering
\fbox {
\begin{subfigure}{\MnistFigureWidth\textwidth}
  \centering
  \includegraphics[width=\linewidth]{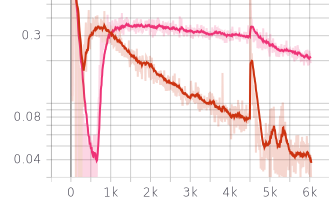}
  	\caption{selector-loss}
 \vspace*{4mm}
\end{subfigure}
}

\fbox {
	\begin{subfigure}{\MnistFigureWidth\textwidth}
	  \centering
  \includegraphics[width=.99\linewidth]{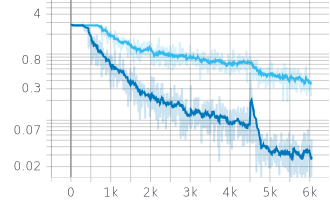}
	  	  \caption{operator-loss-train}
\vspace*{4mm}
	\end{subfigure}

	\begin{subfigure}{\MnistFigureWidth\textwidth}
	  \centering
  \includegraphics[width=.99\linewidth]{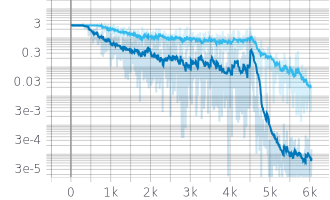}
	  	  \caption{operator-loss-train(${\pmb m}_{opt}$)}
\vspace*{4mm}
	\end{subfigure}

\begin{subfigure}{\MnistFigureWidth\textwidth}
	  \centering
  \includegraphics[width=.99\linewidth]{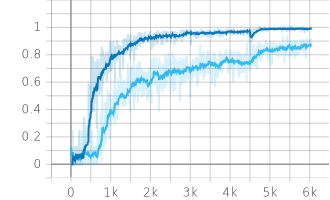}
	  	  \caption{operator-ACC-train}
\vspace*{4mm}
	\end{subfigure}
}

\fbox {
	
	\begin{subfigure}{\MnistFigureWidth\textwidth}
	  \centering
  \includegraphics[width=.99\linewidth]{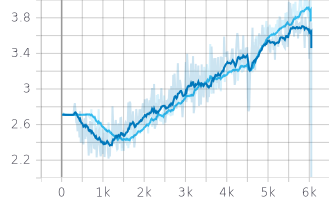}
	  	  \caption{operator-loss-val}
	\end{subfigure}

	\begin{subfigure}{\MnistFigureWidth\textwidth}
	  \centering
  \includegraphics[width=.99\linewidth]{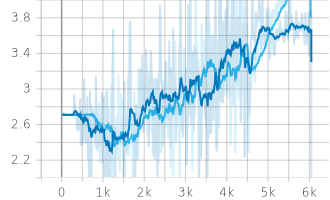}
	  	  \caption{operator-loss-val(${\pmb m}_{opt}$)}
	\end{subfigure}

\begin{subfigure}{\MnistFigureWidth\textwidth}
	  \centering
  \includegraphics[width=.99\linewidth]{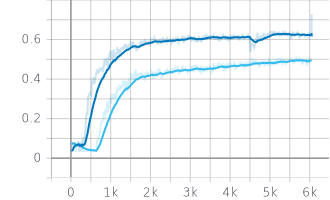}
	  	  \caption{operator-ACC-val}
	\end{subfigure}
}

\caption{\textbf{Yale Benchmark Dataset}. Evolution of the operator and the selector losses in Phase II ($d=784, s=10, 30$). The \texttt{x-axis} corresponds to the number of batches and \texttt{y-axis} refers to the loss statistics of 5 folds. The light/dark colors correspond to $s=10$/$s=30$, respectively.
(a) The selector loss.
(b) The operator loss on the training set.
(c) The operator loss on the training set with ${\pmb m}_{opt}$ only.
(d) The classification accuracy evaluated on the training set.
(e) The operator loss on the validation set.
(f) The operator loss on the validation set with ${\pmb m}_{opt}$ only.
(g) The classification accuracy evaluated on the validation set.
Note that Phase II starts when the operator net has been trained for 4,500 batches in Phase I. The spike at the maximum batch (6k) in (e)-(g) correspond to the results evaluated on the test set with the trained operator net upon the completion of the alternate learning.}
\label{fig:yale_training}
\end{figure}

Next, Figure \ref{fig:training_mnist} illustrates the learning behavior of the operator and the selector in terms of losses in Phase II on the MNIST benchmark subset, a \emph{binary classification} task.
It is seen from Figure \ref{fig:training_mnist}(a) that the evaluation of the averaged selector loss of 5 folds has a reduction trend as the number of batches is increased although the averaged loss not longer drops monotonically.
The sharp selector loss increase at around 10k batches is typical and reflects the nature of our stochastic local search procedure in tackling the combinatorial optimization issue. The sharp increase is likely caused by the fact that the optimal mask identified leads to the sharp operator loss reduction and the selector net did not have such training examples before this moment. This analysis is manifested by all the results at round 10k batches shown in other plots in Figure \ref{fig:training_mnist}.
As evident in Figures \ref{fig:training_mnist}(b) and   \ref{fig:training_mnist}(c), the averaged operator loss on training and further decreases in general. By using an alternative performance index, we also show the averaged classification accuracy measured on the training set in Figure \ref{fig:training_mnist}(d), which allows one to see the learning performance vividly. Likewise, we illustrate the averaged operator loss and accuracy on the validation set in Figures \ref{fig:training_mnist}(e)-(g). Once again, we can see our alternate learning algorithm works very well. Once again, the operator validation loss evaluated with ${\pmb m}_{opt}$ only provides the solid evidence for early stopping. In general, the learning behavior on this binary classification dataset very much resembles that on the nonlinear regression dataset (c.f. Figure \ref{fig:synth_training}). After the alternate learning is completed, we can evaluate the performance of the trained operator net on the test set in the same manner.  To show the test performance, we depict the averaged loss evaluated on the test set with all the optimal mask candidates and the optimal mask as well as the accuracy based on the optimal mask at the maximum batch in Figures \ref{fig:training_mnist}(e)-(g). Interestingly, it is seen in Figures \ref{fig:training_mnist}(e)-(g) that the test performance is significantly better than the validation performance in terms of both the losses and the accuracy. This suggests that our alternate learning algorithm yields the favorable generalization performance on this benchmark dataset.

Finally, Figure \ref{fig:yale_training} shows the learning behavior of the operator and the selector in terms of losses in Phase II on the Yale benchmark dataset, a \emph{multiclass classification} task. For this facial image dataset, we employ a convolutional neural network described in Table \ref{tab:architectures} to carry out the operator net. To understand the learning behavior better, we compare the situations of the alternate learning for different subset sizes, $s=10$ and $s=30$.  It is observed from Figure \ref{fig:yale_training}(a) that the averaged selector loss for different subset sizes behaves quite differently. For $s=10$, the selector loss sharply decreases at the first few hundred batches then sharply increases. The limited amount of information carried in 10 out of 1024 features may be accountable for this phenomenon. In contrast, the evolution of selector loss for $s=30$ is similar to that shown in Figures \ref{fig:synth_training}(a) and \ref{fig:training_mnist}(a). Figures \ref{fig:yale_training}(b)-(d) suggest that the averaged operator loss for different subset sizes keeps decreasing and the accuracy remains increasing on the training set. In contrast, the trend of the averaged operator loss for different subset sizes increases on the validation set after 1.5k batches as shown in Figures \ref{fig:yale_training}(e) and (f). This looks like a typical overfitting scenario. As seen in Figure \ref{fig:yale_training}(g), however, the averaged classification accuracy on the validation set generally keeps increasing regardless of different subset sizes. Furthermore, for $s=30$, the averaged operator test losses and the test accuracy  shown in Figures  \ref{fig:yale_training}(e)-(g) (at 6k batches) also provide the evidence for the good generalization performance. Surprisingly, the  alternate learning behavior on this benchmark dataset contradicts or is inconsistent with the normal behavior of a learning system.  While we do not fully understand such learning behavior, our preliminary analysis implies that the this phenomenon could be caused by the \emph{covariant shift} nature of this facial image dataset and limited training data. In the Yale dataset, the images of an individual subject correspond to different facial expressions. Since there are only limited training examples and the selector learning is constrained by the operator training performance, the stochastic local search in Phase II.A may have to do a lot of exploration in order to find out the ``genuine'' optimal subset (mask). This can be observed by the fluctuated operator validation loss as shown in Figures  \ref{fig:yale_training}(e) and (f). Thanks to our stochastic exploration-exploitation strategy, some sub-optimal subsets may still direct the learning towards the learning performance at an acceptance level.

In summary, we exhibit typical yet different learning behavior of our dual-net architecture trained by the alternate learning algorithm in Figures \ref{fig:synth_training}-\ref{fig:yale_training}. In most of the situations including the one reported in Section \ref{sect:enhance-promoter} and others not reported here, we can use the operator validation loss evaluated with the optimal mask only for early stopping. In some occasion, however, we encounter some ``strange'' learning behavior, as exemplified in Figure \ref{fig:yale_training}. In such occasion, we might have to use the validation classification accuracy (or validation MSE in regression) for early stopping. Thus, we are going to investigate such ``strange" learning behavior in our ongoing work.

\subsection{Detailed Information on Benchmark Data}

In this section, we provide the detailed information on 4 benchmark datasets described in the main text.
In our comparative study, we choose 4 challenging benchmark datasets for feature selection evaluation. As reported in \cite{chen2017CCM}, the state-of-the-art feature selection methods including those latest strong ones do not perform well on the following datasets.

\noindent
\textbf{Glass dataset}\footnote{[online]: \url{https://archive.ics.uci.edu/ml/machine-learning-databases/glass/}}
The Glass is a famous UCI benchmark dataset for a task of predicting a type of glass
based on its chemical composition. Glass dataset contains usually 9 chemical features and the ID for each instance that is normally not treated as a feature. In the experimental setting of CCM \cite{chen2017CCM}, they treated the ID as a new feature so that 10 features are used in their experiments. Due to the fact that the instances in the data file are arranged in an non-shuffled manner according to their class labels, the ID feature turns up to be one of the most important features so that CCM and other strong feature selection methods yields very high accuracy, e.g., CCM achieves 86\% on average \cite{chen2017CCM}. In our experiment, we follow this setting so that our approach yields 90\%+ accuracy (c.f. Figure 4 in the main text). Without the ID feature, however, all the methods including ours yield considerably lower accuracies although ours still outperforms those methods used in our comparative study. In the 5-fold cross validation, the accuracy of our approach drops to the levels of 75\%-80\%, quite close to the known top accuracy of 80\% on the OpenML platform\footnote{[online]: \url{https://www.openml.org/t/3815}}.

\noindent
\textbf{Vowel Dataset}\footnote{[online]: \url{https://www.openml.org/d/307}}
The Vowel is yet another famous UCI benchmark dataset for predicting English vowels from acoustic features. Following the same setting used in CCM \cite{chen2017CCM}, we use a newer version of this dataset so that we can make a fair comparison to those feature selection methods used in our comparative study.

\noindent
\textbf{TOX-171 dataset}\footnote{[online]: \url{http://featureselection.asu.edu/old/datasets.php}}
The TOX-171 is a biological microarray dataset with only 43 instances/class but 5,784 features. The nature of this dataset makes a deep learning model very prone to overfitting. As shown in Figure 4 in the main text, our approach does not outperform the CCM in general, which reveals the limitation of our approach.

\noindent
\textbf{Yale Dataset}\footnote{[online]: \url{http://www.cad.zju.edu.cn/home/dengcai/Data/FaceData.html}}
The Yale is a well-known facial image benchmark dataset. There are 15 individual subjects and 11 images of different facial expressions, e.g., wink, happy and sad, were collected from each individual. When this dataset is used for face recognition, a random split of this dataset could lead to a certain degree of covariant shift; the instances in training and validation/test sets may be subject to different distribution but their distributions conditional on the label are same. This causes a difficulty for all learning models without covariant shift adaptation.

\begin{figure}[H]
\centering

\begin{subfigure}{\MnistFigureWidth\textwidth}
  \centering
  \includegraphics[width=\linewidth]{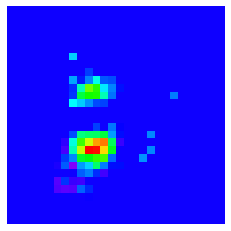}
\end{subfigure}
\begin{subfigure}{\MnistFigureWidth\textwidth}
  \centering
  \includegraphics[width=\linewidth]{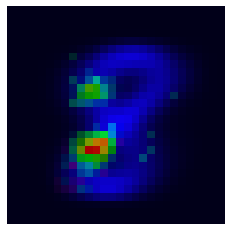}
\end{subfigure}
\begin{subfigure}{\MnistFigureWidth\textwidth}
  \centering
  \includegraphics[width=\linewidth]{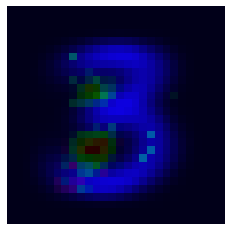}
\end{subfigure}

\begin{subfigure}{\MnistFigureWidth\textwidth}
  \centering
  \includegraphics[width=\linewidth]{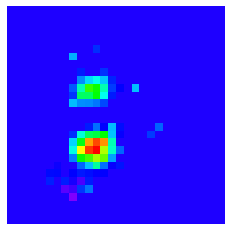}
\end{subfigure}
\begin{subfigure}{\MnistFigureWidth\textwidth}
  \centering
  \includegraphics[width=\linewidth]{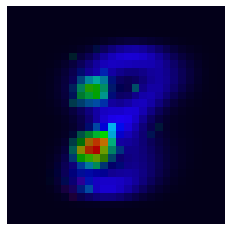}
\end{subfigure}
\begin{subfigure}{\MnistFigureWidth\textwidth}
  \centering
  \includegraphics[width=\linewidth]{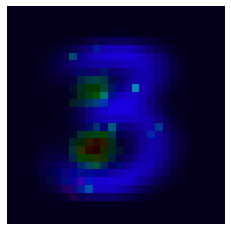}
\end{subfigure}

\begin{subfigure}{\MnistFigureWidth\textwidth}
  \centering
  \includegraphics[width=\linewidth]{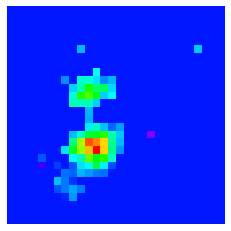}
\end{subfigure}
\begin{subfigure}{\MnistFigureWidth\textwidth}
  \centering
  \includegraphics[width=\linewidth]{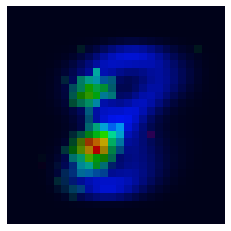}
\end{subfigure}
\begin{subfigure}{\MnistFigureWidth\textwidth}
  \centering
  \includegraphics[width=\linewidth]{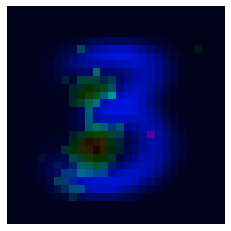}
\end{subfigure}

\begin{subfigure}{\MnistFigureWidth\textwidth}
  \centering
  \includegraphics[width=\linewidth]{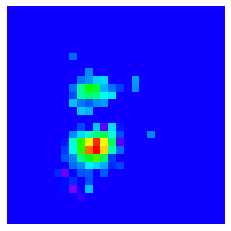}
\end{subfigure}
\begin{subfigure}{\MnistFigureWidth\textwidth}
  \centering
  \includegraphics[width=\linewidth]{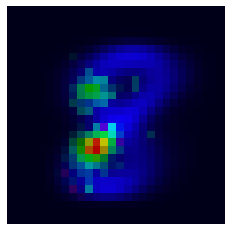}
\end{subfigure}
\begin{subfigure}{\MnistFigureWidth\textwidth}
  \centering
  \includegraphics[width=\linewidth]{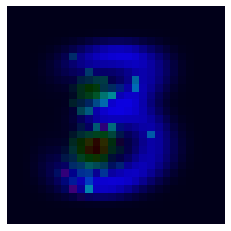}
\end{subfigure}

\begin{subfigure}{\MnistFigureWidth\textwidth}
  \centering
  \includegraphics[width=\linewidth]{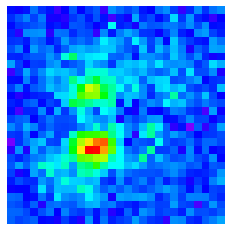}
\end{subfigure}
\begin{subfigure}{\MnistFigureWidth\textwidth}
  \centering
  \includegraphics[width=\linewidth]{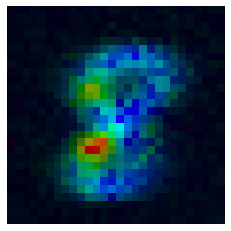}
\end{subfigure}
\begin{subfigure}{\MnistFigureWidth\textwidth}
  \centering
  \includegraphics[width=\linewidth]{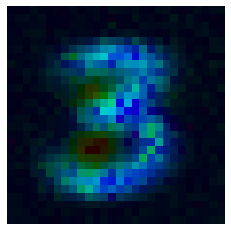}
\end{subfigure}

\caption{{\bf MNIST Subset}. Feature importance maps ($d=784, s=85$) generated with the method described in Section \ref{sect:visual-FIM}. From top to bottom, first 4 rows correspond to feature importance maps achieved from different folds. The bottom row is the full feature importance map corresponding the feature importance map shown in Figure 4 of the main text.  }
\label{fig:mnist_other_folds}
\end{figure}

\subsection{Feature Importance Map}
\label{sect:visual-FIM}

Due to the limited space in the main text, we demonstrate only one feature importance map on the MNIST subset. Below, we show more feature importance maps achieved from other folds on the MNIST subset and those yielded by our approach on the Yale dataset.

To obtain the superimposed feature importance maps on the background image (the mean of raw images), we apply a method as follows. A blank image is first created in the HSV (hue-saturation-value) colour format. The hue used  in [0,270] range corresponds to the importance, and the saturation is set to 1.0 to encode the mean background image from the dataset. Due to the feature importance ranking (FIR) scores are normalised, no negative FIR scores are shown to ensure unselected features have the background color.

Figure \ref{fig:mnist_other_folds} shows different feature importance maps achieved from other 4 folds. As our FIR approach described in Section 3.4 of the main text measure the FIR scores based on the input gradient, it can achieve the input gradient for all the features regardless of whether a feature is selected or not. To this end, we can generate a full feature importance map as well. It is observed from \ref{fig:mnist_other_folds} that the feature importance maps achieved from different folds are very much consistent and the full feature importance map provides a clearer picture in terms of explainablity/interpretability.

\begin{figure}[H]
\centering

\begin{subfigure}{0.195\textwidth}
  \centering
  \includegraphics[angle=270,origin=c,width=\linewidth]{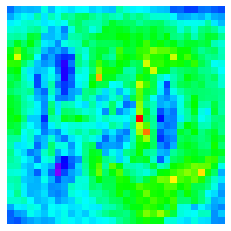}
\end{subfigure}
\begin{subfigure}{0.195\textwidth}
  \centering
  \includegraphics[angle=270,origin=c,width=\linewidth]{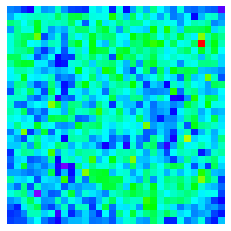}
\end{subfigure}
\begin{subfigure}{0.195\textwidth}
  \centering
  \includegraphics[angle=270,origin=c,width=\linewidth]{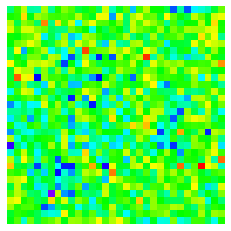}
\end{subfigure}
\begin{subfigure}{0.195\textwidth}
  \centering
  \includegraphics[angle=270,origin=c,width=\linewidth]{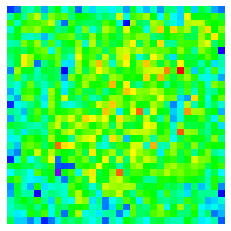}
\end{subfigure}
\begin{subfigure}{0.195\textwidth}
  \centering
  \includegraphics[angle=270,origin=c,width=\linewidth]{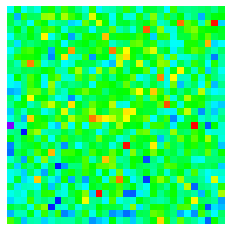}
\end{subfigure}

\begin{subfigure}{0.195\textwidth}
  \centering
  \includegraphics[angle=270,origin=c,width=\linewidth]{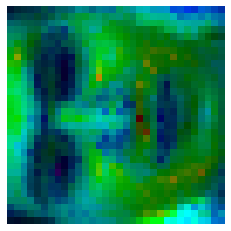}
\end{subfigure}
\begin{subfigure}{0.195\textwidth}
  \centering
  \includegraphics[angle=270,origin=c,width=\linewidth]{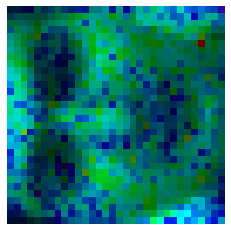}
\end{subfigure}
\begin{subfigure}{0.195\textwidth}
  \centering
  \includegraphics[angle=270,origin=c,width=\linewidth]{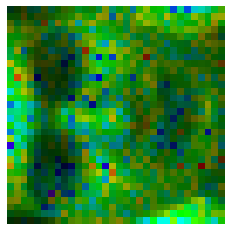}
\end{subfigure}
\begin{subfigure}{0.195\textwidth}
  \centering
  \includegraphics[angle=270,origin=c,width=\linewidth]{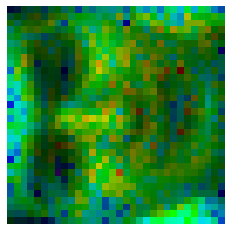}
\end{subfigure}
\begin{subfigure}{0.195\textwidth}
  \centering
  \includegraphics[angle=270,origin=c,width=\linewidth]{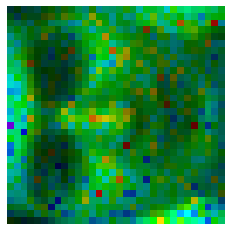}
\end{subfigure}
\caption{{\bf Yale Dataset}. Feature importance maps achieved with different subset mask sizes: $s=10,30,50,70,90$ (from left to right) out of $d=32 \times 32$. The second row corresponds to the images generated by superimposing the feature importance maps to the background image, i.e., mean face image averaged on 11 images collected from an individual.}
\label{fig:yale_faces}
\end{figure}

As Yale is a facial image dataset, we can also illustrate the feature importance maps in Figure \ref{fig:yale_faces} for visual inspection. The visual inspection reveals that increasing the mask size $s$ results in less clear visual representation of feature importance. In comparison, the best performed mask size of 30 clearly selects several meaningful yet discriminative features, e.g., pixels near lip, nose and eyes, and ranks their importance properly as shown in the 2nd column in Figure \ref{fig:yale_faces}. As this dataset has limited instances (7 training examples/class on average), we reckon that the use of large subset mask size is likely to cause the overfitting as revealed by their feature importance maps shown in Figure \ref{fig:yale_faces}.

\begin{figure}[ht]
\centering
\fbox {
\begin{subfigure}{\MnistFigureWidth\textwidth}
  \centering
  \includegraphics[width=\linewidth]{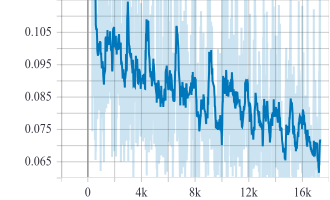}
  	\caption{selector-loss}
  \vspace*{4mm}
\end{subfigure}

}

\fbox {
\centering
	\begin{subfigure}{\MnistFigureWidth\textwidth}
	  \centering
  \includegraphics[width=1.02\linewidth]{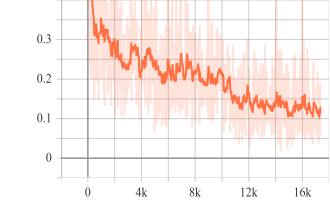}
	  	  \caption{operator-loss-train}
\vspace*{4mm}
	\end{subfigure}

	\begin{subfigure}{\MnistFigureWidth\textwidth}
	  \centering
  \includegraphics[width=1.02\linewidth]{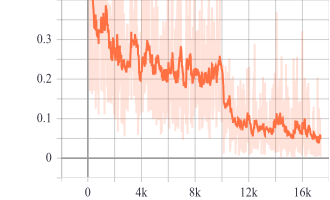}
	  	  \caption{operator-loss-train(${\pmb m}_{opt}$)}
\vspace*{4mm}
	\end{subfigure}

\hspace*{1mm}

\begin{subfigure}{\MnistFigureWidth\textwidth}
	  \centering
  \includegraphics[width=1.01\linewidth]{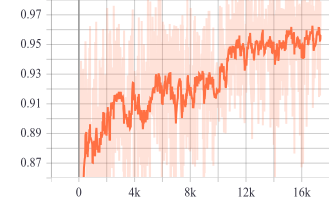}
	  	  \caption{operator-ACC-train}
\vspace*{4mm}
	\end{subfigure}

}

\fbox {

	\begin{subfigure}{\MnistFigureWidth\textwidth}
	  \centering
  \includegraphics[width=.99\linewidth]{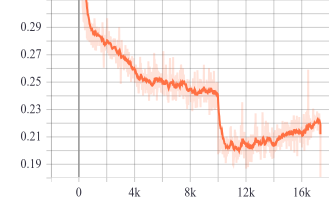}
	  	  \caption{operator-loss-val}
	\end{subfigure}

	\begin{subfigure}{\MnistFigureWidth\textwidth}
	  \centering
  \includegraphics[width=.99\linewidth]{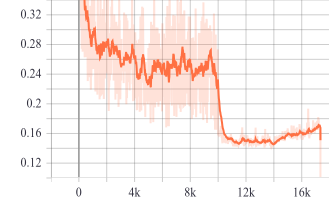}
	  	  \caption{operator-loss-val(${\pmb m}_{opt}$)}
	\end{subfigure}

	\begin{subfigure}{\MnistFigureWidth\textwidth}
	  \centering
  \includegraphics[width=.99\linewidth]{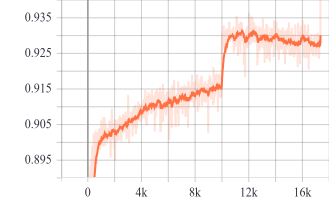}
	  	  \caption{operator-ACC-val}
	\end{subfigure}

}

\caption{\textbf{Enhancer-Promoter Dataset}. Evolution of the operator and the selector losses in Phase II ($d=102, s=35$). The \texttt{x-axis} corresponds to the number of batches and \texttt{y-axis} refers to the loss statistics of 5 folds.
(a) The selector loss.
(b) The operator loss on the training set.
(c) The operator loss on the training set with ${\pmb m}_{opt}$ only.
(d) The classification accuracy evaluated on the training set.
(e) The operator loss on the validation set.
(f) The operator loss on the validation set with ${\pmb m}_{opt}$ only.
(g) The classification accuracy evaluated on the validation set.
Note that Phase II starts when the operator net has been trained for 10,000 batches in Phase I. The spike at the maximum batch  in (e)-(g) correspond to the results evaluated on the test set with the trained operator net upon the completion of the alternate learning.}
\label{fig:training}
\end{figure}

\begin{figure}[ht]
\centering
  \includegraphics[width=.52\linewidth]{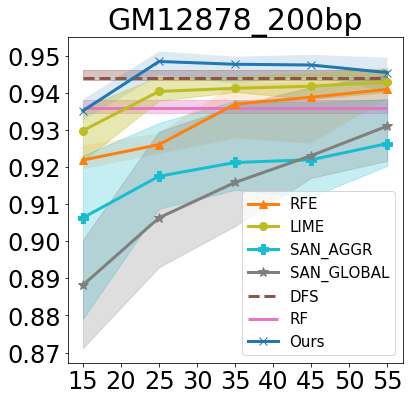}

\caption{Classification accuracies yielded by different methods
 on the Enhancer-Promoter dataset: cell line GM12878 (200dp). The shadowed regions refer to the performance range between the minimum and the maximum accuracies on 5 folds.While DFS and RF yield only one result with all the 102 features, other methods produce the results at different subset sizes for $s=15, 25, 35, 45, 55$.  }
\label{fig:enhance_promoter_perf}
\end{figure}

\section{More Results on Real-World Data}
\label{sect:enhance-promoter}

In this section, we report more results on the enhance-promoter dataset and the UCI gene expression cancer RNA-Seq data set.

\subsection{Results on Enhance-Promoter Dataset}

To evaluate our approach on real-world data, we adopt the GM12878 cell line (a lymphoblastoid cell line) dataset \cite{Li2016DFS}.
This is the dataset for which the \emph{deep feature selection} (DFS) method \cite{Li2016DFS} is especially proposed. Therefore, we follow their setting by using only annotated DNA regions of GM12878 cell line (200 bp).

In the original dataset, there are 7 classes and 102 features, each class has 3,000 instances apart from one that has only 2878 instances. The 7 classes are \emph{\textbf{active promoter, active enhancer}, active exon, inactive promoter, inactive enhancer, inactive exon}  and \emph{unknown regions}. The main interest in medicine is classifying the function of DNA sequences into enhancer, promoter and background since non-coding gene regulatory enhancers are essential to transcription in mammalian cells \cite{anderson2014enhancer}.

Following the suggestion in \citep{anderson2014enhancer, Li2016DFS}, we merge inactive enhancers, inactive promoters, active exons, and unknown regions into a background class. Thus, we have a 3-class imbalanced dataset as the background class has roughly 5 times more instances than other two classes: active promoter and active enhancer. We follow a preprocessing method consisting of two steps: 1) making the dataset balanced by down-sampling so that each of 3 classes has 2878 samples; 2) overcoming the natural skewness of biological outcome by taking the logarithm on the input. To avoid the zero-value issue in logarithm, we append a small positive number to each feature, $x \leftarrow x + 0.01$, in our experiments. Note that step 2) is not described in the DFS paper but we believe that this is important for such a data distribution.

It is also worth clarifying that we see some discrepancies between the data presented in the authors' repository\footnote{[online]: \url{https://github.com/yifeng-li/DECRES/tree/master/data}\label{fn:en-pro}}
and their article \cite{Li2016DFS}. Two main differences include 1) 102 features in the repository but 92 features stated in the article; 2) at least 2,878 instance/class in the repository but only 2,156 features mentioned in the article. While we use the dataset in their repository, we have done our best by keeping all the settings suggested in their article for a fair comparison in our experiments.

Due to the limited space in the main text, we only report the result of our approach for a subset mask size, $s=35$. Here, we report more results of our approach and other methods used in comparative study on this dataset.

We first illustrate the learning behavior of our dual-net model on this real-world dataset in Figure \ref{fig:training}. As shown in Figure \ref{fig:training}(a), the averaged selector loss has the typical behavior as described in Section \ref{sect:behavior}. The loss fluctuation and the loss reduction trend in the loss evolution vividly exhibit how the stochastic local search strategy works in finding out optimal subset masks. As shown in Figures \ref{fig:training}(b)-\ref{fig:training}(g), the learning progresses steadily as evident in the evolution of the averaged operator losses and the averaged classification accuracies on the training and the validation sets. Also it is clearly seen in Figures \ref{fig:training}(b)-\ref{fig:training}(g), the overfitting occurs once the optimal subset mask is identified at around 10.5k batches. Once again, this observation provides the solid evidence to support for early stopping with the operator training/validation losses measured on the optimal subset masks (${\pmb m}_{opt}$). Furthermore, it is also seen in Figures \ref{fig:training}(e)-\ref{fig:training}(g) (at the maximum batches) that the averaged test losses and the averaged classification accuracy yielded by the trained operator net on 5 folds are superior to their counterparts on the validation set. Once again, this suggests that our alternate learning leads to the favorable generalization performance although an earlier stopping may yield a better accuracy.

Apart from the comparative study specified in the main text, we have conducted the further experiments on this dataset for a comparison to two recent state-of-the-art methods
\cite{linden2019explanationbb,skrlj2020fiesan}
 that obtain the populationwise FIR by aggregating the instancewise FIR. In \cite{linden2019explanationbb}, the global aggregation method workable on this dataset is the homogeneity-weighted importance, which is the same as the global LIME importance proposed in \cite{ribeiro2016should}. In our experiment, we use an MLP of
the architecture: 102-300-200-50-3 and \texttt{n\_samples=500} to achieve the LIME importance on the validation set \cite{ribeiro2016should}. In the SAN
\cite{skrlj2020fiesan}, the populationwise FIR is obtained via either
instance-level aggregation (SAN\_AGGR) or global attention layer (SAN\_GLOBAL). In our experiments, we use the same settings suggested by the authors \cite{skrlj2020fiesan}\footnote{We do this experiment with the authors' code: \url{https://gitlab.com/skblaz/attentionrank}.} with the following hyperparameters: $k=1$, \texttt{p\_{dropout}=20\%}, \texttt{epochs=32}, \texttt{batch\_size=32}, \texttt{n\_1\_1=128} (number of hidden neurons in SAN). In terms of feature selection, both methods fall into the filtering category. Therefore, we employ an MLP of the architecture: s-300-200-50-3 to be a classifier trained on those selected feature subsets of $s=5,25,35,45,55$, respectively, for this 3-class classification task.

\begin{figure}[h]
\centering
	\begin{subfigure}{1.0\textwidth}
	  \centering
  \includegraphics[width=\linewidth]{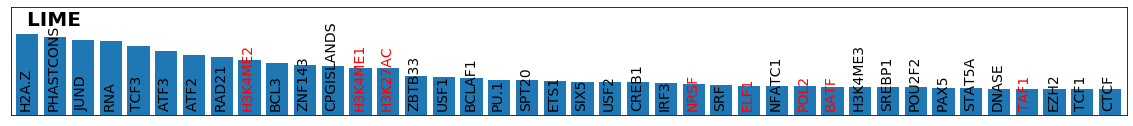}
	\end{subfigure}
		\begin{subfigure}{1.0\textwidth}
	  \centering
  \includegraphics[width=\linewidth]{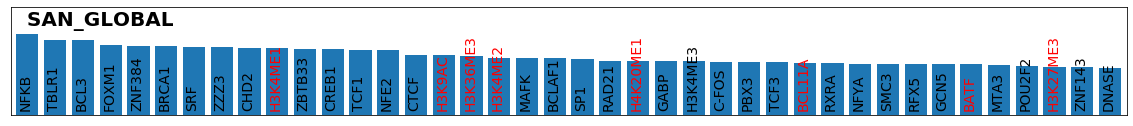}
	\end{subfigure}
		\begin{subfigure}{1.0\textwidth}
	  \centering
  \includegraphics[width=\linewidth]{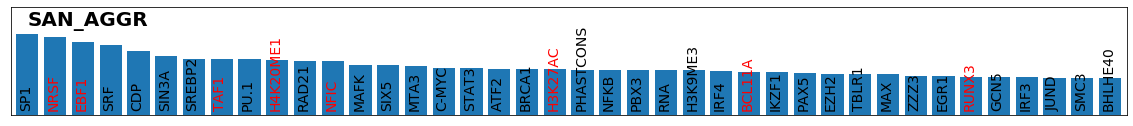}
	\end{subfigure}
	\begin{subfigure}{1.0\textwidth}
	  \centering
  \includegraphics[width=\linewidth]{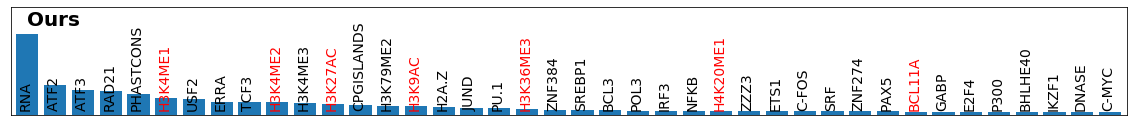}
	\end{subfigure}
	
\caption{Feature importance ranking (FIR) scores yielded by LIME, SAN and ours for top 55 features ($s=55$) on the Enhancer-Promoter dataset: GM12878 Cell line (200 bp).}
\label{fig:enhance-promoter1}
\end{figure}

We report the accuracies yielded by 6 different methods for different subset sizes. As shown in Figure \ref{fig:enhance_promoter_perf}, it is evident that our approach yields slightly better accuracies than the DFS \cite{Li2016DFS}, a method especially proposed for this biological dataset, when the subset size of selected features is larger than 15. Also, our approach outperforms RFE \cite{guyon2002RFE}, a state-of-the-art feature selection method especially effective on gene selection for cancer classification, and RF \cite{breiman2001RF}, a famous off-the-shelf ensemble learning model. In contrast, it is evident from  Figure \ref{fig:enhance_promoter_perf} that ours along with DFS also outperforms those methods of using the aggregation of instancewise FIR to obtain populationwise one at all different subset sizes ranging from 15 to 55.

For feature importance ranking (FIR), we show the FIR scores yielded by two instancewise aggregation-based methods and ours for $s=55$ in Figure \ref{fig:enhance-promoter1}. It is observed that two methods and ours yield different FIR results  and different settings in the SAN also results in different FIR for top 55 features. Due to a lack of the ground-truth, it is difficult to draw an affirmative conclusion but the experimental results suggest that the populationwise FIE is an extremely challenging problem for real-word data.

\begin{figure}[H]
\centering
	\begin{subfigure}{1.0\textwidth}
	  \centering
  \includegraphics[width=\linewidth]{figure/CellLine/dfs_dfs_new_200.png}
	\end{subfigure}
		\begin{subfigure}{1.0\textwidth}
	  \centering
  \includegraphics[width=\linewidth]{figure/CellLine/dfs_rf_new_200.png}
	\end{subfigure}
		\begin{subfigure}{1.0\textwidth}
	  \centering
  \includegraphics[width=\linewidth]{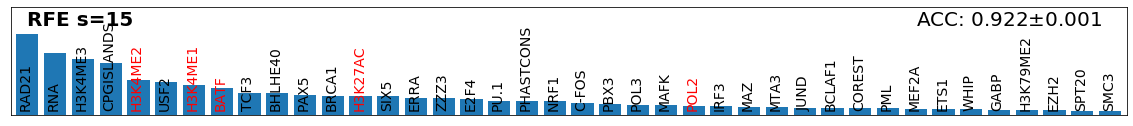}
	\end{subfigure}
			\begin{subfigure}{1.0\textwidth}
	  \centering
  \includegraphics[width=\linewidth]{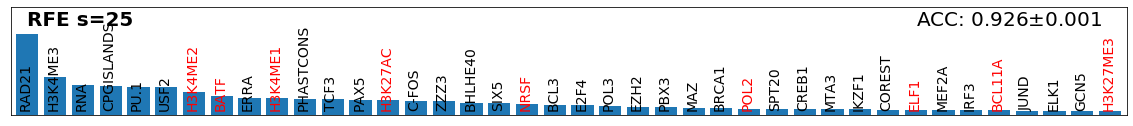}
	\end{subfigure}
			\begin{subfigure}{1.0\textwidth}
	  \centering
  \includegraphics[width=\linewidth]{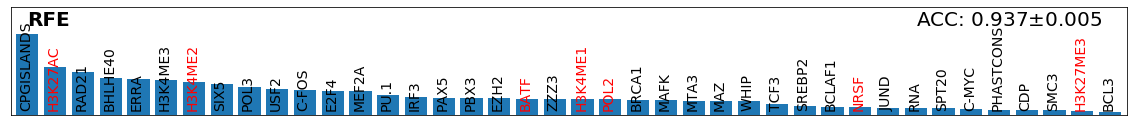}
	\end{subfigure}
			\begin{subfigure}{1.0\textwidth}
	  \centering
  \includegraphics[width=\linewidth]{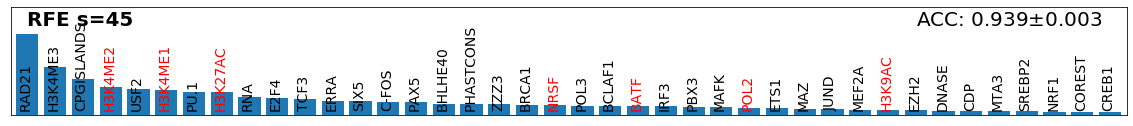}
	\end{subfigure}
			\begin{subfigure}{1.0\textwidth}
	  \centering
  \includegraphics[width=\linewidth]{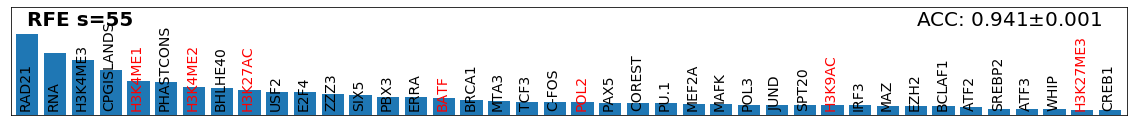}
	\end{subfigure}
		\begin{subfigure}{1.0\textwidth}
	  \centering
  \includegraphics[width=\linewidth]{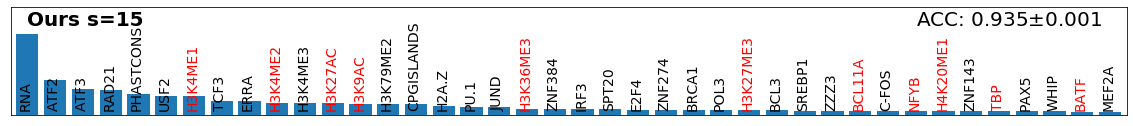}
	\end{subfigure}
			\begin{subfigure}{1.0\textwidth}
	  \centering
  \includegraphics[width=\linewidth]{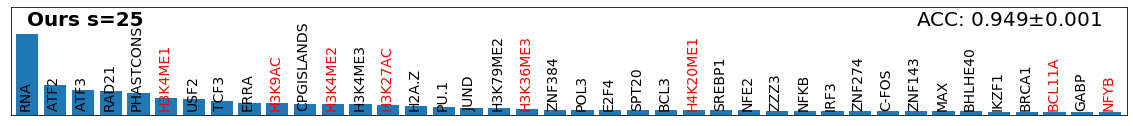}
	\end{subfigure}
			\begin{subfigure}{1.0\textwidth}
	  \centering
  \includegraphics[width=\linewidth]{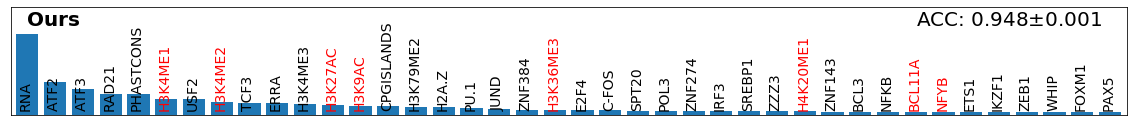}
	\end{subfigure}
			\begin{subfigure}{1.0\textwidth}
	  \centering
  \includegraphics[width=\linewidth]{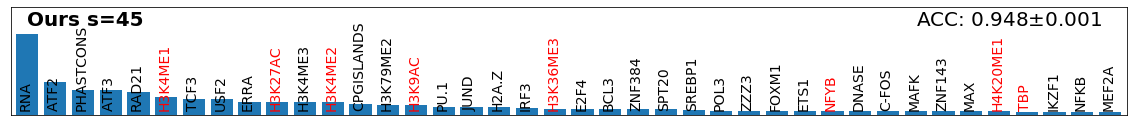}
	\end{subfigure}
			\begin{subfigure}{1.0\textwidth}
	  \centering
  \includegraphics[width=\linewidth]{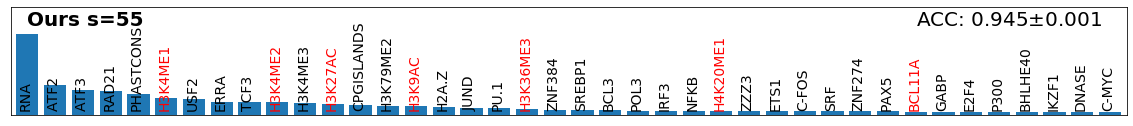}
	\end{subfigure}
	
\caption{Accuracy and feature importance ranking (FIR) scores yielded by different methods on the Enhancer-Promoter dataset: GM12878 Cell line (200 bp). While DFS and RF yield only one result with all the 102 features, RFE and ours produce the results at different subset sizes for $s=15, 25, 35, 45, 55$. Note that the results yielded RFE and ours for $\textcolor{red}{s=35}$ above are not specified deliberately with the subset size to indicate that those have been reported in the main text.}
\label{fig:enhance-promoter}
\end{figure}

We further show the FIR scores for top-40 features produced by DFS, RF, RFE and ours in Figure \ref{fig:enhance-promoter}. The FIR scores of the RFE and the RF are generated based on the RFE feature importance estimator \cite{guyon2002RFE} and the out-of-bag errors \cite{breiman2001RF}. The FIR score of the DFS is achieved based the magnitude of shrunk weights between input and the first hidden layer introduced in the DFS method  \cite{Li2016DFS}. From Figure \ref{fig:enhance-promoter}, it is observed that our approach yields the relatively consistent FIR results when different subset sizes are used given the fact that the importance ranking order of top features only varies for one or two. Also, our approach is the only one to constantly rank ``RNA'' and two important genes ``ATF2"'' and ``ATF3'' among the most important features regardless of feature subset sizes.  In comparison, the DFS also selects those two genes but does not rank them on the top. On the other hand, the RFE chooses other genes, ``RAD21'', "``PGISLANDS'' and ``H3K4ME3'', as the most important features irrespective of feature subset sizes. It is also be seen in Figure \ref{fig:enhance-promoter} that the DFS and ours, two deep learning models rank the importance of features similarly but differently from the RFE and the RF that yield similar FIR scores. Our experimental results on this real-world dataset suggest that deep learning models may lead to different results from the existing state-of-the-art and off-the-shelf machine learning models for FIR. Thus, learning models of different types should be considered simultaneously and their results can be fused at the discretion of domain experts in such real-world applications.

To evaluate the efficiency, we record the averaging training time on this dataset in terms of 5-fold cross-validation. our experiments are conducted on a server of the specification and the environment: Intel Core i7-5930K CPU (3.50GHz), NVIDIA GeForce GTX TITAN X GPU, 64 GB RAM and CentOS 7. In summary, our algorithm takes around 1,100 sec while RF, SAN, DFS, LIME and RFE take around 2.5, 35, 90, 540 and 1,700 sec, respectively. The dual-net architecture along with the alternate learning is responsible for a high computational load in our approach,

\newpage

\subsection{Results on RNA-seq Data}

Apart from the comparative study on the Enhancer-Promoter dataset, we have further applied our approach to the UCI gene expression cancer RNA-Seq data set\footnote{[online]: \url{https://archive.ics.uci.edu/ml/datasets/gene+expression+cancer+RNA-Seq}}, to evaluate our approach on a dataset of many features.

The gene expression cancer RNA-Seq dataset is part of The Cancer Genome Atlas Pan-cancer
Analysis Project. The original data set is maintained by the
cancer genome atlas pan-cancer analysis project.
Gene expression data are composed of DNA microarray and RNA-seq data. Therefore,  microarray data analysis facilitates the clarification of
biological mechanisms and development of drugs toward a more predictable future. In comparison to hybridization-based microarray technology,
RNA-seq has a larger range of expression levels and contains more information.
RNA-Seq is a random extraction of gene expression of patients with
five different types of tumors including BRCA (breast), KIRC
(kidney), COAD (colon), LUAD (lung) and PRAD (prostate).
 The dataset contains 801 samples, each of which has 20,531 biological features or genes. The data set is imbalanced and there are 300, 146, 78, 141 and 136 samples for BRCA, KIRC, COAD, LUAD and PRAD, respectively.

Unlike other methods, e.g., \cite{GARCIADIAZ20201916}, we do not pre-process the imbalanced data in our experiment apart from removal of 267 constant features. In other words, we use only 20,264 features in each sample to train our dual-net model. All the data are standardised with zero mean and unit standard deviation. The dataset is randomly split into two subsets, training and test, where there are 600 and 201 samples in the training and the test subset. The four-fold cross-validation working on the training subset are used for parameter estimate and hyperparameter tuning for our dual-net model. To make a fair comparison to the best performer on this dataset as reported in  \cite{GARCIADIAZ20201916}, we use the identical setting by using $s=49$ in our experiment. The information on the dual-net architecture and optimal hyperparameter values used in this experiment is provided in Tables \ref{tab:architectures} and \ref{tab:hyperparameters}, respectively.

\begin{table}[h]
\begin{center}
\caption{Accuracy yielded by different methods on RNA-seq dataset (adapted from Table 7 in \cite{GARCIADIAZ20201916}).}
\makebox[\textwidth][c]{
\begin{tabular}{|c|c|c|c|c|c|c|}
\hline
  Method & \# Samples & \# Features & \# Classes & \# Selected Features & \# Classifiers & Accuracy \\
\hline
\cite{DING2005}  & 96 & 4026 & 9 & $<$60 & 1 & 0.9730\\
\cite{MAHATA2007775}  & 62 & 6000 & 2  & 15 & 1 & 0.9677\\
\cite{LIU20102763}  & 97 & 24481  & 2 & 7& 7 & 0.9381\\
\cite{LIU20102763}  & 102 & 12600 & 2 & 4 & 7 &0.9706 \\
\cite{BEST2015666}  & 175 & 1072 & 2 & - & 110 & 0.9500 \\
\cite{PIAO201739}  & 215 & 1047 & 4 & - & 20 & 0.9860 \\
\cite{SAYGILI2018}  & 569 & 32& 2& 24 & 1& 0.9877 \\
\hline
\cite{GARCIADIAZ20201916}  & 801 & 20531 & 5 & 49 & 5 & 0.9881 \\
\textbf{Ours}  & \textbf{801}& \textbf{20531} & \textbf{5} & \textbf{49} & \textbf{1} &\textbf{0.9938} \\

\hline
\end{tabular}
}
\label{tab:rna_seq_table}
\end{center}
\end{table}

Table \ref{tab:rna_seq_table} shows the existing results of several feature selection methods \cite{GARCIADIAZ20201916,DING2005,MAHATA2007775,LIU20102763,BEST2015666,PIAO201739,SAYGILI2018}
on this dataset with various settings although most of the existing methods do not work on the entire dataset. From Table \ref{tab:rna_seq_table}, it is clearly seen that, under the same settings, our approach outperforms the best performer on this dataset in literature.
The experimental result on this nontrivial real-world dataset demonstrates that our approach works well for a data set of many features as long as there are enough training examples required by deep learning. Thus, we firmly believe that our approach will be applicable to a large data set of many features, e.g., images where there are millions of pixels. We shall look into the scalability of our approach in our ongoing work.

In summary, our experimental results manifest that leveraged with deep learning, our approach  outperforms a number of state-the-art FIR and feature selection methods on two biological datasets. This suggests that our approach would be a strong candidate for feature selection and feature importance ranking in real-world biological applications.

\newpage

\section{Pseudo Code}
\label{sect:pseudo}

In this section, we describe the implementation of our alternate learning algorithm used to train our proposed dual-net neural architecture for feature importance ranking underlying feature selection. The pseudo code\footnote{Our source code and all the related information regarding the experimental settings are available online: \url{https://github.com/maksym33/FeatureImportanceDL}.} in
Algorithm \ref{algo:main-steps} carries out the alternate learning algorithm as described in Sect. 3.3 of the main text. The pseudo code in Algorithm \ref{algo:opt-mask} implements a subroutine used in line 10 of Algorithm \ref{algo:main-steps} to generate an optimal feature subset in the current step as described in Phase II.A (c.f. Sect. 3.3 in the main text). In Algorithm \ref{algo:main-steps}, lines 1-7 corresponds to Phase I, lines 9-12 carry out Phase II.A and lines 13-18 implement Phase II.B. Phase II.A and II.B alternate until the early stopping condition is satisfied as implemented by the loop from line 8 to line 23.

\vspace*{5mm}

\begin{algorithm}[ht]
\caption{Alternate Learning Algorithm}
\label{algo:main-steps}
\begin{algorithmic}[1]
      \REQUIRE loss function of operator net, $ l(\pmb x \otimes \pmb m, \pmb y; \theta)$, selector net, $f_{S}(\varphi,\pmb m)$
    \REQUIRE feature set/subset size $d$ and $s$, fraction of random masks $f$, perturbation factor $s_p$
    \REQUIRE number of optimal subset candidates $|\mathcal{M}'|$, number of batches $E_1$ in Phase I.
    \REQUIRE mask weight vector used in the weighted selector loss $\pmb w_{S}$ of $|\mathcal{M}'|$ elements.
   	\notrainglecomment{
	
	    }
%
	\FOR{$e \gets 0$ to $E_1$} 
        \STATE ${\cal M}_1'= \big \{ {\pmb m}_i| {\pmb m}_i = {\rm Random}({\cal M}, s) \big \}_{i=1}^{|{\cal M}'|}$ \COMMENT{create a random batch of masks}

	\STATE ${\cal L}_O ({\cal D}, {\cal M}_1'; \theta) = \frac 1 {|{\cal M}'||{\cal D}|}
\sum_{\pmb m \in {\cal M}'} \sum_{(\pmb x, \pmb y) \in {\cal D}}  l(\pmb x \otimes \pmb m,\pmb y; \theta)$ \COMMENT {calculate operator loss}
	\STATE $\theta'' \triangleq \theta' - \eta \nabla_{\theta}
{\cal L}_O ({\cal D}, {\cal M}_1'; \theta)|_{\theta=\theta'}$ \COMMENT{update the parameters in operator}
 	
	\STATE ${\cal L}_S ({\cal M}'; \varphi ) =  \frac 1 {2|{\cal M}'|} \sum_{\pmb m \in {\cal M}'}
\Big ( f_S(\varphi; \pmb m) - \frac 1 {|{\cal D}|} \sum_{({\pmb x}, {\pmb y}) \in {\cal D}}
l(\pmb x \otimes \pmb m, \pmb y; \theta) \Big )^2$ \COMMENT{calculate MSE loss of selector net}
	\STATE $\varphi'' \triangleq \varphi' - \eta \nabla_{\varphi}
{\cal L}_S ({\cal M}_1'; \varphi)|_{\varphi=\varphi'}$
 	\COMMENT{update the weights in selector}
	\ENDFOR 
%
%
%
	\FOR{$t \gets 0$ to $\inf$}
  \STATE ${\cal M}_{t+1,1}'= \big \{ {\pmb m}_i| {\pmb m}_i = {\rm Random}({\cal M}, s) \big \}_{i=1}^{(1-f)|{\cal M}'|}$ \COMMENT{create a random batch of masks}

	\STATE $m_{t+1,opt} \Leftarrow \mathrm{\textit{generateOptimalMask}} (f_{SN}(\varphi_t))$ \COMMENT{implemented in Algorithm \ref{algo:opt-mask} }
	\STATE ${\cal M}_{t+1,2}' = \{{\pmb m}_{t,best}\} \cup
 \{{\pmb m}_{t+1,opt}\} \cup \big \{{\pmb m}_i|{\pmb m}_i={\rm Perturb}({\pmb m}_{t+1,opt},s_p) \big \}_{i=1}^{f|{\cal M}'|-2}$ \COMMENT{collect the best mask from last step, the current optimal mask and those perturbed optimal masks}
	\STATE ${\cal M}_{t+1}' = {\cal M}_{t+1,1}' \cup {\cal M}_{t+1,2}'$ \COMMENT{form new subset candidates for operator}
	\STATE ${\cal L}_O ({\cal D}, {\cal M}_1'; \theta) = \frac 1 {|{\cal M}'||{\cal D}|}
\sum_{\pmb m \in {\cal M}'} \sum_{(\pmb x, \pmb y) \in {\cal D}}  l(\pmb x \otimes \pmb m, \pmb y; \theta_t)$ \COMMENT {calculate operator loss}
	\STATE $\theta_{t+1} \triangleq \theta_t - \eta \nabla_{\theta}
{\cal L}_O ({\cal D}, {\cal M}_1'; \theta)|_{\theta=\theta}$ \COMMENT{update parameters in operator}
 	
	\STATE ${\cal L}_S ({\cal M}'; \varphi ) =  \frac 1 {2|{\cal M}'|} \sum_{i=0}^{|{\cal M}'|}
 {w_{S,i}} \Big ( f_S(\varphi; \pmb m) - \frac 1 {|{\cal D}|} \sum_{({\pmb x}, {\pmb y}) \in {\cal D}}
l(\pmb x \otimes \pmb m, \pmb y; \theta) \Big )^2$ \COMMENT{calculate the weighted MSE loss of selector}
	\STATE $\varphi_{t+1} \triangleq \varphi_t - \eta \nabla_{\varphi}
{\cal L}_S ({\cal M}_1'; \varphi)|_{\varphi=\varphi_t}$ \COMMENT{update the parameters in selector}
	\STATE $\pmb m_{t+1,best} = \text{argmin}_{\pmb m \in {\cal M}_{t+1}'}\Big( \sum_{({\pmb x}, {\pmb y}) \in {\cal D}}
l(\pmb x \otimes \pmb m, \pmb y; \theta)\Big)$ \COMMENT{record the best performed mask}
	\STATE $\mathcal{L}_{t,m_{opt}} \leftarrow \big(\sum_{({\pmb x}, {\pmb y}) \in {\cal D}} l(\pmb x \otimes \pmb m, \pmb y; \theta)  \big)[(1-f)+1]$ \COMMENT{record loss of the ${\pmb m}_{opt}$, $(1-f)+1$ should be its index}
	\IF{$checkEarlyStopping(\mathcal{L}_{t,m_{opt}})$}
		\STATE{$\theta_t = restoreBestWeights()$} \COMMENT{stopping condition is met}
		\STATE{\bf break}
	\ENDIF
	\ENDFOR
\end{algorithmic}
\end{algorithm}

\begin{algorithm}[t]
\caption{Generation of Optimal Feature Subset}
\label{algo:opt-mask}
\begin{algorithmic}[1]
    \REQUIRE selector net $f_{S}(\varphi,\pmb{m})$
    \REQUIRE input feature set size $d$, selected subset size  $s$
	\STATE $\pmb m_0 \leftarrow (\frac{1}{2},\frac{1}{2},...,\frac{1}{2}) $
    \STATE $\pmb \delta_{m_0} = \frac{\partial f_{S}(\varphi,\pmb m)}{\partial \pmb m}|_{\pmb m=\pmb m_0}$ \COMMENT{calculate input gradient}
    \STATE $\pmb m_{opt} \leftarrow (0,0,...,0) $
    \STATE $(\pmb i_{unmasked},\pmb i_{masked} )\leftarrow \mathrm{argsort}(\pmb \delta_{m_0}))$ \COMMENT{determine indexes with 1s (unmasked, top $s$ biggest gradients) and 0s (masked, the smallest gradients)}
    \STATE $\pmb m_{opt}[\pmb i_{unmasked}] \leftarrow (1,1,...,1) $\COMMENT{complete creating ${\pmb m}_{opt}$}
        \STATE $\pmb \delta_{m_{opt}} = \frac{\partial f_{SN}(\varphi,\pmb m)}{\partial \pmb m}|_{\pmb m=\pmb m_{opt}}$ \COMMENT{reclalculate the gradient at $\pmb m=\pmb m_{opt}$} \label{step:m_opt_validation}
    \STATE $i_{min} \leftarrow \mathrm{argmin}(\pmb \delta_{m_{opt}}[i_{unmasked}])$ \COMMENT{index of minimum unmasked gradient}
    \STATE $i_{max} \leftarrow \mathrm{argmax}(\pmb \delta_{m_{opt}}[i_{masked}])$ \COMMENT{index of maximum masked gradient}

    \STATE $\pmb i_{neg} \leftarrow \mathrm{argwhere}(\pmb \delta_{m_{opt}}[i_{unmasked}] <0)$\COMMENT{create a set of unmasked indices that have negative gradients}
    \FOR{$i$ in $\pmb i_{neg}$}
    	\STATE $\pmb m'_{opt} \leftarrow \pmb m_{opt} $ \label{step:m_opt_step_1}\COMMENT{\textbf{Validation step 1}}
	    \STATE $\pmb m'_{opt}[i] \leftarrow 0 $ \COMMENT{mask the negative, previously unmasked, input}
		\STATE $\pmb m'_{opt}[i_{max}] \leftarrow 1 $ \COMMENT{unmask the biggest (gradient-wise), previously masked input}
		\IF{$f_{S}(\varphi,\pmb m''_{opt})<f_{S}(\varphi,\pmb m_{opt})$}
    	\STATE $\pmb m_{opt}\leftarrow \pmb m'_{opt}$  \COMMENT{replace $\pmb m_{opt}$ and restart the validation}
    	\STATE recalcualte $\pmb i_{unmasked}$ and $\pmb i_{masked}$
    	\STATE goto step \ref{step:m_opt_validation}
		\ENDIF
    \ENDFOR
    \STATE $\pmb m''_{opt} \leftarrow \pmb m_{opt} $ \label{step:m_opt_step_2}\COMMENT{\textbf{Validation step 2}}
	\STATE $\pmb m''_{opt}[i_{min}] \leftarrow 0 $ \COMMENT{mask the smallest (gradient-wise), previously unmasked, input}
	\STATE $\pmb m''_{opt}[i_{max}] \leftarrow 1 $ \COMMENT{unmask the biggest (gradient-wise), previously masked input}
    \IF{$f_{S}(\varphi,\pmb m''_{opt})<f_{SN}(\varphi,\pmb m_{opt})$}
    	\STATE $\pmb m_{opt}\leftarrow \pmb m''_{opt}$  \COMMENT{replace $\pmb m_{opt}$ and restart the validation}
    	\STATE recalculate $\pmb i_{masked}$ and $\pmb i_{masked}$
    	\STATE goto step \ref{step:m_opt_validation}
    \ENDIF

\end{algorithmic}
\end{algorithm}

\newpage

\small
\bibliographystyle{unsrt}
\bibliography{mybib}